\documentclass[10pt,letterpaper]{article}
\usepackage[T1]{fontenc}
\usepackage{lmodern}
\usepackage[margin=1in]{geometry}
\usepackage[authoryear,round]{natbib}
\setcitestyle{citesep={;},aysep={,},yysep={;}}
\usepackage{graphicx}
\usepackage{booktabs}
\usepackage{amsmath,amssymb}
\usepackage{multirow}
\usepackage{xcolor}
\usepackage{url}
\usepackage[hidelinks]{hyperref}
\usepackage{caption}
\usepackage{subcaption}
\usepackage[section]{placeins}
\usepackage{enumitem}
\usepackage{needspace}
\usepackage{tikz}
\usetikzlibrary{positioning,arrows.meta}
\setlist{itemsep=2pt,parsep=0pt,topsep=3pt,partopsep=0pt}

\hypersetup{
  pdftitle={Compiling Known Rules into Compact Neural Agents: Weight-Space Programming on Unseen Games},
  pdfauthor={Jens Lehmann}
}
\title{\sffamily\bfseries Compiling Known Rules into Compact Neural Agents:\\
Weight-Space Programming on Unseen Games}
\author{Jens Lehmann\thanks{Work done outside of Amazon.}\\
{\small\itshape Amazon, TU Dresden}}
\date{}

\begin{document}
\maketitle

\begin{abstract}
Frontier language models have recently been reported to write the weights of neural networks
directly, without training: weight-space programming (WSP). If they can do so for tasks they have
never seen, a model's understanding of a task could be compiled into a compact network that runs
without the model and might be refined by training. We test this on 20 privately authored, never
published game families in the style of ARC-AGI-3. A language model, the writer, receives a game's
simulator source without the held-out levels and must produce one network that plays held-out
levels, renderings and start states. When the architecture is fixed (a recurrent convolutional
network with 57k parameters), the best writer's networks solve 56\% of episodes on held-out levels
with feedback, but the four strongest writers' Python programs solve 97--100\%, and the gap grows with game
difficulty. Learning this architecture's weights from the levels shown to the writers, instead of
writing them, transfers far worse (at most 11\%), and learned weights fail even on those levels
once the rendering changes, which costs written weights at most 8 points. Writers read the source
while learners see only those levels, so this compares information access. When writers design their own network in a small
tensor language, an exploratory condition, these four writers solve 68--90\%. On closer inspection, these networks are programs in network form: 93\% of their
parameters are exactly zero, 90\% of the rest are small integers or powers of two, and NumPy
programs with the same loop rules and caps solve 91\% against the networks' 79\%. In these games,
and under a protocol that forbids fitting and search, WSP thus amounts to writing a restricted class of programs, not networks of the kind training
produces. Nor does the neural format bring its main hoped-for benefit, training. The best writer's
architectures, trained from standard initialisation within our budget, solve 1\%. Written weights in the fixed network are also
fragile: weight noise the size of PPO's first update removes most of their success, as does the update itself. Asked for noise-robust weights, two of three writers tolerate about $10^4$ times more noise, but PPO's
first update on them is 10--100 times larger and still removes most of their success. With standard PPO on these binary-success games, writing and RL fine-tuning do
not combine.
\end{abstract}

\section{Introduction}
\label{sec:intro}

Frontier language models can solve tasks they have never seen by writing programs, for example
programs that play new games. A stranger claim is that they can also write the \emph{weights} of a
neural network directly: construct networks for simple tasks, read what small networks compute,
and repair failures by editing parameters~\citep{takehi2026sparks,singh2026wsp}. If such
\emph{weight-space programming} (WSP) works, a model's reasoning about a task could be compiled into
a compact neural agent. Such an agent would run without the model, with small, fixed computation per
step instead of a large model's long and variable reasoning, and, being differentiable, might be
refined by reinforcement learning (RL). We find that written networks do not behave like trained
ones: they are programs in network form, their architectures fail when trained from scratch within
our budget, and written fixed-actor weights do not survive standard PPO fine-tuning.

We are interested in three research questions about WSP. Do written networks generalise to
instances their writer never saw? Do they add anything over an ordinary program written by the same
model, for example because they can be trained further or run more cheaply? Do they add anything
over learning the same network from the instances the writer was shown? Existing demonstrations use
public tasks (Breakout, Go, arithmetic), where recall of familiar strategies cannot be excluded, and
rarely compare against programs or against learning in the same architecture.

We study these questions on 20 privately authored game families in the style of
ARC-AGI-3~\citep{arcprize2026arcagi3}. Games are a demanding test: a player must read the frame,
remember what it has done and act over many steps, and new levels show whether it generalises. We
report the 25 public ARC-AGI-3 games as a separate, possibly familiar tier. In our setup, a language
model, the writer, receives a game's simulator source with solvers and held-out levels removed. The
remaining \emph{support levels} are the ones it can study and get feedback on. It must produce one
actor that plays held-out levels, renderings and start states. Nine writer models from OpenAI and
Anthropic either set the weights of a fixed 57k-parameter recurrent convolutional actor, design and
set a network of their own in a small tensor language (\emph{free architecture}, FreeArch), or write a
Python player, each one-shot or with feedback. We compare against learned weights in the same actor
and against simple replay and fixed-action baselines.

\needspace{4\baselineskip}
\paragraph{Findings.}
\begin{enumerate}
\item \emph{In the fixed actor, written weights generalise, but programs are far better.} Main
writers' actors solve 33--56\% of held-out-level episodes with feedback, far above the baselines
(${\le}4\%$), while their Python players solve 97--100\%, and the gap grows with difficulty. Where a
program can solve the task, as here, writing it is the better approach. Tasks
without a compact program may differ.
\item \emph{Learned weights in the fixed actor transfer poorly, and written weights are
appearance-invariant.} Learning the actor's weights from the support levels, instead of writing
them, transfers at most 11\%, although the actor can fit held-out episodes when trained on them.
The learned actors fail even on the levels they were trained on once the rendering changes: a
held-out renderer template alone costs them 34--79 points and written weights 1--8.
Writers read the source and learners see 15--30 level starts per family, so this compares
information access more than learning.
\item \emph{With a free architecture, writers write programs in network form.} In networks of their
own design, an exploratory condition, main writers solve 68--90\%, mostly owing to the free topology. The networks are
bounded, branch-free tensor programs, and NumPy programs with the same loop rules and caps do better
(91\% against 79\%). Writers had to construct weights by design, without fitting or search, which
favours such networks. The ability is concentrated in the strongest models: the other current and
the previous-generation writers reach only 14--48\%.
\item \emph{The ordering holds on the public games.} On the 25 public ARC-AGI-3 games, which are
harder and possibly familiar, programs (43--75\%) beat free architectures (13--42\%), which beat
fixed-actor weights (${\le}8\%$).
\item \emph{Written weights and standard PPO do not combine.} In the fixed actor, written weights
do not survive PPO fine-tuning: weight noise the size of PPO's first update removes most of their
success, and PPO removes it, as does a random update of the same size. The writers' architectures also
fail when trained from scratch within our budget (1\% for the best writer). So the main hoped-for
benefit of the neural format is not realised here, and written networks are also slower per step
than Python players.
\end{enumerate}

\paragraph{Contributions.} The findings above are our main contribution. They rest on (i) a
controlled protocol for weight-space programming on unseen interactive tasks: audited
mechanics-only sources, held-out final levels, certified perturbed starts, closed-loop controls and
complete request accounting for the main study; (ii) a multi-writer comparison of written weights,
in a fixed actor and in free architectures, with programs by the same models, learned weights in
the same actor, and controls that separate the language, the topology and the operators; and
(iii) tests of training from scratch, refinement by RL and cost, with weight-perturbation controls.

\section{Related work}
\label{sec:related}

\paragraph{Weight-space programming.} \citet{takehi2026sparks} report that a frontier model can
tell from raw weights what a small network was trained to do, write MinAtar and $9\times9$ Go
policy networks from the rules, and repair networks by direct edits. Their games are public.
\citet{singh2026wsp} had a coding agent set every weight of small transformers for ten algorithmic
tasks, and document evaluation leakage, forbidden fitting and lookup-table solutions, which motivate
our sandbox and held-out design. A later version also writes language classifiers and brain
encoding models~\citep{singh2026iwsp}. Follow-up work edits small language models for reasoning
benchmarks~\citep{hu2026exploring} and audits small clinical networks~\citep{nusrat2026auditing}.
On the reading side, \citet{ahmed2026can} find that zero-shot weight reading relies on local
arithmetic and dataset priors more than on tracing computation.
We differ from this work in four ways. Our games are private, so writers cannot recall solutions,
and every network is scored on held-out levels, renderings and start states. Each written network
is compared with programs by the same model and with weights learned in the same architecture. We
test whether written networks can be trained further, with RL and perturbation controls. And we
cover nine writers across two model generations, in a fixed architecture and in architectures the
writers design.

\paragraph{Rules and programs compiled into networks.} KBANN maps a propositional domain theory to
the topology and initial weights of a network~\citep{towell1994knowledge}, finite-state automata can
be programmed stably into recurrent networks~\citep{omlin1996constructing}, and neural-symbolic work
generalises such translations~\citep{garcez2002neural,besold2017neural}. A minimal convolutional
network can run the Game of Life exactly, yet networks of that size rarely learn
it~\citep{springer2021hard}, and gradient descent in differentiable interpreters rarely finds the
programs that discrete search finds~\citep{gaunt2016terpret}. RASP and Tracr compile programs into transformer
weights~\citep{weiss2021thinking,lindner2023tracr}, with extensions to learnable~\citep{friedman2023learning},
looped~\citep{shaw2025alta,giannou2023looped} and embedded~\citep{weng2024mastering} programs. There,
a person or an exact compiler translates a formal rule language. Our writers start from simulator
source, must also invent the policy, and give no correctness guarantee: we measure correctness on
held-out instances.

\paragraph{Generating weights.} HyperNetworks~\citep{ha2017hypernetworks},
G.pt~\citep{peebles2022learning}, p-diff~\citep{wang2024neural}, neural
functionals~\citep{navon2023equivariant,zhou2023permutation} and
Text-to-LoRA~\citep{charakorn2025text} learn a mapping into weight space by gradient descent,
mostly from collections of trained networks~\citep[see also][]{wang2026weightspace}. Our writers
receive no trained networks or gradients, only the task's source and feedback.

\paragraph{Algorithmic reasoning in networks.} CLRS~\citep{velickovic2022clrs,velickovic2021neural},
value iteration networks~\citep{tamar2016value} and weight-tied recurrent networks that extrapolate
by iterating longer~\citep{schwarzschild2021can,bansal2022end} ask whether gradient descent finds
algorithmic solutions. We ask whether a writer can set them directly. Our fixed actor contains such
a weight-tied recurrent block, iterated 128 times per step, so a writer can use it to run an
iterative algorithm, for example to spread information across the grid.

\paragraph{Language models writing code that acts.} Language models write robot
policies~\citep{liang2023code}, skill libraries~\citep{wang2024voyager}, reward
functions~\citep{ma2024eureka}, world models~\citep{tang2024worldcoder,dainese2024generating},
evolved programs~\citep{romeraparedes2024mathematical}, programmatic
policies~\citep{liu2025synthesizing} and game-playing algorithms~\citep{li2026modular}. On
ARC~\citep{chollet2019measure,chollet2025arcagi2}, sampled and verified programs are among the
strongest approaches~\citep{wang2024hypothesis,li2025combining,chollet2024arcprize}. ARC-AGI-3
turns such tasks into games that require exploration and rule
discovery~\citep{arcprize2026arcagi3}, where coding agents build programmatic world models from
interaction~\citep{lehmann2026tycho} and training-free graph exploration is a strong
baseline~\citep{rudakov2025graph}. Our writers instead read the game's source, so we measure how
well they compile known rules. The weights conditions ask for the program as the parameters of a
network, so that no language-model call, planner or search remains at run time.

\paragraph{RL fine-tuning and distillation.} Fine-tuning can forget capabilities on states that are
not visited early~\citep{wolczyk2024finetuning}, which JSRL~\citep{uchendu2023jump} and residual
policy learning~\citep{silver2018residual,johannink2019residual} avoid by keeping the prior policy
fixed. Our distillation baseline builds on policy
distillation~\citep{rusu2015policy,czarnecki2019distilling,hinton2015distilling} and
DAgger~\citep{ross2011reduction}.

\section{Experimental setup}
\label{sec:setup}

\subsection{Games and held-out scope}
\label{sec:games}
We study generalisation within a game: given a game's rules, the writer must produce one actor
that plays new levels, layouts, renderings and states of that game. Games the writer was never told
about are left for future work. ``Unseen'' refers to the writer: the private games were never
published, although familiar ideas such as path finding, parity and matching are known to the
models.

\begin{figure}[t]
\centering
\begin{minipage}[c]{0.645\linewidth}\centering
\resizebox{\linewidth}{!}{\begin{tikzpicture}[font=\scriptsize, node distance=1.2mm and 5mm,
  box/.style={draw=black!50, rounded corners=2pt, align=center, inner sep=2.5pt, fill=#1},
  box/.default=white, arr/.style={-{Stealth[length=1.6mm]}, semithick},
  sq/.style={draw=black!60, minimum size=2.4mm, inner sep=0pt}]
\node[box=black!4, minimum width=4.0cm, minimum height=2.15cm] (game) {};
\node[anchor=north, align=center] at ([yshift=-0.6mm]game.north) {\textbf{Game family}: mechanics-only\\source and support levels};
\foreach \i in {1,...,8} {
  \pgfmathsetmacro{\held}{mod(\i,2)==0}
  \ifdim\held pt=1pt \def\c{black!35}\else\def\c{white}\fi
  \node[sq, fill=\c] at ([xshift={1.72cm+(\i-1)*2.8mm}, yshift=-1.0mm]game.west) {};}
\node[anchor=west] at ([xshift=1.2mm, yshift=-1.0mm]game.west) {levels 1--8};
\foreach \i in {0,...,9} {
  \ifnum\i>4 \def\c{black!35}\else\def\c{white}\fi
  \node[sq, fill=\c] at ([xshift={1.72cm+\i*2.2mm}, yshift=-5.2mm]game.west) {};}
\node[anchor=west] at ([xshift=1.2mm, yshift=-5.2mm]game.west) {templates};
\node[anchor=south, text=black!70] at ([yshift=0.4mm]game.south) {grey: held out};
\node[box=orange!18, right=of game, text width=1.25cm] (wr) {\textbf{Writer}\\(language\\model)};
\node[box=blue!8, right=6mm of wr, text width=3.5cm, yshift=6.6mm] (w) {\textbf{W1/W2}: weights of the fixed actor};
\node[box=yellow!18, below=of w, text width=3.5cm] (fa) {\textbf{FreeArch W1/W2}: own network + weights};
\node[box=red!8, below=of fa, text width=3.5cm] (p) {\textbf{P1/P2}: Python player};
\node[box=green!12, below=2.4mm of p, text width=3.5cm] (lb) {\textbf{Learned}: same actor trained on support levels};
\node[box=black!4, right=6mm of fa, text width=2.6cm, yshift=-1.5mm] (ev) {\textbf{Evaluation} on held-out
  levels, templates and perturbed starts};
\draw[arr] (game.east |- wr) -- (wr);
\foreach \t in {w,fa,p} \draw[arr] (wr.east) -- (\t.west);
\foreach \t in {w,fa,p,lb} \draw[arr] (\t.east) -- (ev.west);
\draw[arr] (game.south) |- (lb.west);
\draw[arr, dashed] ([xshift=-3mm]w.north) -- ++(0,2.2mm) -| node[pos=0.27, above]
  {feedback on support starts (W2, FreeArch W2, P2)} (wr.north);
\end{tikzpicture}}
\end{minipage}\hfill
\begin{minipage}[c]{0.335\linewidth}
\includegraphics[width=\linewidth]{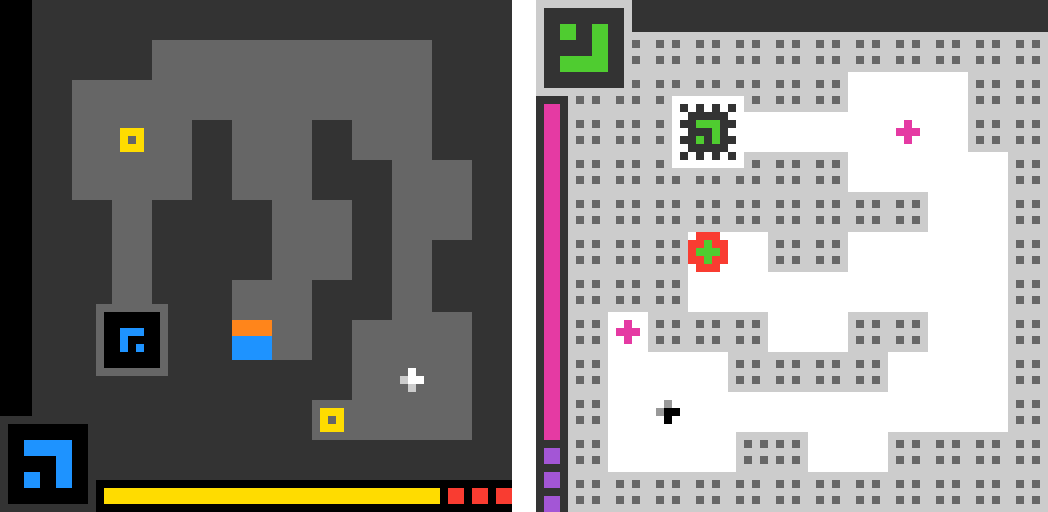}
\end{minipage}
\caption{Left: setup. A writer receives a game's mechanics-only source and returns weights for the
fixed actor, a network of its own design, or a Python player, one-shot (1) or with feedback (2).
Learned baselines train the fixed actor on the support levels. Right: one state of the public game
\texttt{ls20}~\citep{arcprize2026arcagi3} as published and under a template we constructed for this
figure. The board and HUD are turned, player, collectibles, walls and the key frame are drawn in
other styles, and colours are permuted, which changes 91\% of the cells while the mechanics stay the
same. The private templates vary geometry, object and HUD style and palette in this way.}
\label{fig:setup}
\end{figure}

\paragraph{Private tier (primary).} 20 ARC-AGI-3-style game families (Table~\ref{tab:families},
Appendix~\ref{app:repro-sets}). An observation is a $64\times64$ grid of 16 colour ids. The actions are
\texttt{RESET}, \texttt{ACTION1}--\texttt{5}, \texttt{ACTION6} (a click at a cell) and
\texttt{ACTION7}. Before any result, the game author assigned each family to a difficulty stratum:
3 easy, 11 medium, 5 hard and 1 very hard. Core computations include object comparison, colour
matching under a budget, maze planning with global phase changes, click localisation with
cascades, constraint covering, timing and hidden state. Each family has 6--12 fixed levels and ten
renderer \emph{templates} that keep the mechanics but vary geometry, object and HUD grammar, and
palette. A held-out template is not an interpolation of the shown ones, and the colour ids that
carry a role change across templates and palettes, so a detector keyed to fixed colour ids cannot
work (Figure~\ref{fig:setup} shows an example, Appendix~\ref{app:repro-sets}). Every family comes with an exact planner that we wrote, a solver that returns a shortest
action sequence completing a level from a given state. It certifies evaluation instances and
supervises or shapes learned baselines, but is never used in the writer or actor. Three of the 20
families (\texttt{ztl}, \texttt{lys}, \texttt{esm}) were used in a preliminary study that informed the
protocol and prompts (Appendix~\ref{app:repro-writers}).

\paragraph{Public tier (secondary).} The 25 public ARC-AGI-3 games are widely discussed,
so writers may know them, and we report them separately. They have no renderer templates, so each game has a single rendering.

\paragraph{Splits and evaluation sets.} For a game with $n$ levels we hold out levels $n, n-2,
\dots$ down to level~2: about half the levels, always including the final one. Templates 0--4 are
shown and 5--9 held out. The primary set, \textbf{Q-lvl}, contains every held-out level under all
ten templates, each from the level's start and from up to five \emph{certified perturbed starts}. A
perturbed start is reached by 1--6 random actions that change the observation. It is kept only if
the planner's direct plan from it replays to completion. We report held-out levels between shown
ones (Q-lvl-mid) and the final level (Q-lvl-final) separately. \textbf{Q-tpl} renders support
levels with held-out templates, and \textbf{S-start} holds the support level starts used for
feedback. On the public tier, each held-out level has its start and up to 20 perturbed starts. An episode
ends on level completion (success), on \texttt{GAME\_OVER} or after $3\times$(certified plan
length)$+10$ actions; \texttt{RESET} restarts the level and counts as an action. All sets are built
mechanically before the main study's first writer call (Appendix~\ref{app:repro-sets}).

\paragraph{Mechanics-only source.} The writer receives the game's source code, the simulator that
applies actions and renders frames, without held-out level data and template parameters (replaced by \texttt{WITHHELD} placeholders), without any solver,
planner or oracle routine or value derived from one, such as optimal move counts, and without
comments that discuss levels or strategy. The redacted simulator reproduces the
original frame for frame on every support level and template (525k steps), and raises an error
when asked to build a held-out level or template, whose data it no longer contains
(Appendix~\ref{app:repro-redaction}). Ablation M+S keeps the solver.

\subsection{The fixed actor}
\label{sec:actor}
All fixed-architecture neural conditions produce the
weights of one recurrent convolutional actor with 57{,}289 float32 parameters
(Appendix~\ref{app:repro-actor}). We designed it before the study to express the computations grid
games need, with operators simple enough to set by hand, not for particular games:
\begin{itemize}[nosep]
\item one-hot colour planes and coordinate channels, for colours and positions;
\item a $5\times5$ and a $3\times3$ convolution, for local patterns and objects;
\item row- and column-wise max broadcasts, for same-row and same-column relations;
\item a weight-tied $3\times3$ block iterated 128 times, for flood fills and distance
fields~\citep{schwarzschild2021can,bansal2022end,tamar2016value};
\item spatial pooling, an MLP with a 64-dimensional episodic memory, an action head and a click map,
for decisions that depend on earlier steps.
\end{itemize}
Decoding is deterministic argmax over available actions. The actor runs alone, without search,
planner or external calls, at 1.32~GMAC per step. Final actor evaluations use exact FP32; the training-time curves in Figure~\ref{fig:stage2}
use TF32 checkpoints. Written networks are numerically brittle: TF32 convolutions change the decisions of 11 of 60 tested written
actors (Appendix~\ref{app:integrity}).

\subsection{Free architecture}
\label{sec:fa-setup}
In the free-architecture condition (FreeArch), the writer designs its own network in a small static
tensor language and supplies all of its weights. We specified the language after the fixed-actor
results were known, so the FreeArch analyses are exploratory. The language offers convolutions,
elementwise arithmetic including multiplication and min/max, row, column and global pooling, dense
layers, weight-tied loops of fixed length and an episodic memory, but no data-dependent control
flow. Inputs are the fixed actor's plus the available-action mask. Caps on graph size, loop
length, arithmetic (1.5~GMAC per step, close to the fixed actor's 1.33) and parameters (150k,
2.6$\times$ the fixed actor's) match the two settings roughly on arithmetic, not on parameters or
wall-clock cost (Appendix~\ref{app:repro-actor}). The prompt shows the fixed actor in the language as an example.

Two controls separate what FreeArch changes, with the same writers, families, runs and feedback. In
\textbf{control A}, the graph must be the fixed actor's own graph written in the language, so only
the weights are free. In \textbf{control B}, any graph is allowed, but only with the fixed actor's
operations. The \textbf{bounded program player} writes NumPy code under FreeArch's loop rules and
caps, with more operations (division, comparisons, \texttt{np.where} and others) and without node or
channel limits (Appendix~\ref{app:repro-actor}). It shows what the weight format costs against a
program with the same limits.

\subsection{Writers and conditions}
\label{sec:conditions}
\paragraph{Writers.} The four \emph{main writers} are GPT-6 Astra, GPT-6.1 Sol, GPT-6.1 Terra and
Claude Opus 5.5. The five \emph{weaker writers} are GPT-6 Luna, Claude Sonnet 5.5, Claude Haiku 5.5
and the previous-generation GPT-5.6 Sol and Claude Opus 5. All run at reasoning effort \texttt{high},
without tools, with the same prompt: the mechanics-only source and the engine base class, the support
start observations, the episode rules and, for the weights conditions, the actor's exact NumPy
reference (Appendices~\ref{app:repro-prompts} and~\ref{app:repro-writers}).

\paragraph{Conditions} (Figure~\ref{fig:setup}).
\begin{itemize}[nosep]
\item \textbf{Written weights}, one-shot (\textbf{W1}) or with feedback (\textbf{W2}): the writer returns
\texttt{build()}, which constructs all tensors of the fixed actor. A sandbox prohibits training,
fitting, searching over weights and simulating the game (Appendix~\ref{app:repro-sandbox}). Feedback
allows up to three resubmissions, each after an outcome table on the support starts and up to three
failure reports (frames, action traces and, for networks, logits and channel maxima).
\item \textbf{FreeArch W1} and \textbf{FreeArch W2}: the same in the free architecture.
\item \textbf{Python players}, one-shot (\textbf{P1}) or with feedback (\textbf{P2}):
\texttt{act(obs, available, memory)}. They control for game understanding.
\item \textbf{P-react}: a one-shot Python player that may not simulate the game or search, with fixed
computation per frame and 50\,ms per step. It is a program control closer to what a network
computes. The restrictions are prompted and audited, not enforced (Appendix~\ref{app:repro-sandbox}).
\end{itemize}
W1, W2, P1 and both FreeArch arms run for all nine writers and the other arms for the main writers,
with three runs per family (Table~\ref{tab:requests}). P-react and the public-tier one-shot arms run
for the seven current writers. Ablations on six families, with one run each, add solver code, remove
the actor's axial broadcasts, cut its tied iterations to 8, or lower the reasoning effort.

\paragraph{Hypotheses.} Four hypotheses about the fixed actor on the private tier, and the
fixed-actor analyses, were written down in a protocol before data collection (excerpt and dated amendments in the supplement):
\begin{itemize}[nosep]
\item \textbf{H1}: the main writers' written weights solve held-out instances well above the replay
and open-loop floors (Section~\ref{sec:eval}).
\item \textbf{H2}: written weights generalise better than weights learned in the same actor from the
same support data, with augmentations that the source shows to preserve the game's rules.
\item \textbf{H3}: weight writing degrades with game difficulty faster than program writing.
\item \textbf{H4}: written actors are closed-loop and do not replay support sequences.
\end{itemize}

\subsection{Learned baselines}
\label{sec:baselines}
All fixed-architecture learned baselines train the identical 57k actor from the same initialisation, on support levels
and templates unless noted (Appendix~\ref{app:repro-baselines}). \textbf{BC-support} is
DAgger~\citep{ross2011reduction} with the planner as teacher. \textbf{BC-aug} adds augmentations
that the source shows to preserve the game's rules: colour permutations, and board symmetries where
the game's own templates use them (a symmetry-only variant runs on 13 families). \textbf{BC-oracle}
also trains on states of the held-out levels and is a capacity check, not a generalisation result.
\textbf{Distill} and \textbf{distill-aug} use a writer's final P2 player as teacher, without and
with augmentation. \textbf{PPO}~\citep{schulman2017proximal} uses a sparse completion reward
($2\times10^6$ steps), and the privileged \textbf{PPO-shaped} adds potential-based
shaping~\citep{ng1999policy} from the planner's distance to go. \textbf{Learned FreeArch networks}
are Astra's FreeArch graphs with re-initialised weights, trained by BC-support and by distillation
for at most 6~h per run.

\subsection{Evaluation and statistics}
\label{sec:eval}
\paragraph{Floors and closed-loop controls.} Two floors measure how much of a game can be solved
without reading the frame. The \emph{replay floor} replays, on every held-out start, the planner's
plan for the support start whose first frame differs from it in the fewest cells. The
\emph{open-loop floor} is the best fixed schedule: one action, or two alternating ones, repeated
until the episode ends. We would have excluded as blind-solvable any family on which a floor reaches 0.5. None does, and no public game scores above 0. For every written actor we measure
success with a constant observation (\emph{zero frame}) or one frozen at the episode's first frame
(\emph{first frame}), success on perturbed starts with \texttt{RESET} disabled, and how often the action changes on counterfactual frames.

\paragraph{Scoring and statistics.} Refusals, truncated replies, invalid artifacts and undelivered
sessions score 0; Table~\ref{tab:prespec} reports delivery, scores without undelivered sessions and
other protocol-specified analyses. The unit of analysis is the family: runs are
averaged within a family, then families are weighted equally (family macro). Q-lvl success is
level-balanced (the mean of per-level success; within 0.02 of the raw episode rate at the reported family-macro cell level). Confidence intervals
come from a hierarchical bootstrap over families, held-out levels and runs; paired differences are
bootstrapped over families. Appendix~\ref{app:integrity} covers numerics and integrity.

\section{Results}
\label{sec:results}

\subsection{Written weights generalise (H1)}
\label{sec:h1}
Table~\ref{tab:main} gives held-out-level success (Q-lvl) for every writer and condition on the 20
private families. GPT-6 Astra's written actors solve 40\% of held-out-level episodes one-shot (W1)
and 56\% with feedback (W2). With feedback, GPT-6.1 Sol reaches 43\%, Claude Opus 5.5 36\% and
GPT-6.1 Terra 33\%. The other current models solve at most 6\% one-shot, and previous-generation
models at most 12\%. Replaying the plan of the nearest support start solves 2.5\%, and
the best repeating action schedule 3.9\%. No family exceeds 0.26 under the first floor or 0.46
under the second. H1 holds in every difficulty stratum except the very hard family (Table~\ref{tab:strata}). Without the preliminary-study families, W2 is 0.27--0.50 and P2 0.96--1.00 (Table~\ref{tab:prelim}).

Every main writer improves with feedback, by 11--21 points and in 15--17 of 20 families. Runs
vary widely: 31 of Astra's 60
feedback runs reach 50\% on held-out levels, and 25 of Sol's. Selection matches part of the feedback
gain: the best of three one-shot runs, chosen by support-start success at a similar request count,
recovers 45--70\% of it (Appendix~\ref{app:results}).

\begin{table}[t]
\centering\small
\caption{Held-out-level success (Q-lvl) on the 20 private families: equal-weight family macro
with 95\% hierarchical-bootstrap CIs (families $\to$ held-out levels $\to$ runs). W = written
weights in the fixed actor, P = Python player; 1 = one-shot, 2 = with up to three feedback rounds.
``W1 deliv.'' is the fraction of W1 sessions that returned a valid artifact (refusals and invalid
artifacts score 0). ``W2 final lvl'' is W2 success on the held-out final level only. P2 was run for
the four main writers. $^\dagger$ previous generation.}
\label{tab:main}
\begin{tabular}{lllllll}
\toprule
Writer & W1 & W2 & P1 & P2 & W1 deliv. & W2 final lvl\\
\midrule
GPT-6 Astra & 0.40 {\scriptsize[0.26,0.54]} & 0.56 {\scriptsize[0.40,0.71]} & 0.95 {\scriptsize[0.86,1.00]} & 1.00 {\scriptsize[1.00,1.00]} & 1.00 & 0.40\\
GPT-6.1 Sol & 0.32 {\scriptsize[0.17,0.49]} & 0.43 {\scriptsize[0.28,0.59]} & 0.96 {\scriptsize[0.91,1.00]} & 1.00 {\scriptsize[0.98,1.00]} & 0.98 & 0.31\\
Claude Opus 5.5 & 0.15 {\scriptsize[0.06,0.26]} & 0.36 {\scriptsize[0.20,0.52]} & 0.85 {\scriptsize[0.73,0.94]} & 0.97 {\scriptsize[0.91,1.00]} & 0.85 & 0.25\\
GPT-6.1 Terra & 0.21 {\scriptsize[0.09,0.36]} & 0.33 {\scriptsize[0.19,0.48]} & 0.97 {\scriptsize[0.92,1.00]} & 0.98 {\scriptsize[0.95,1.00]} & 0.98 & 0.18\\
GPT-6 Luna & 0.05 {\scriptsize[0.01,0.11]} & 0.11 {\scriptsize[0.04,0.19]} & 0.45 {\scriptsize[0.31,0.60]} & -- & 0.70 & 0.02\\
Claude Sonnet 5.5 & 0.06 {\scriptsize[0.02,0.13]} & 0.19 {\scriptsize[0.08,0.33]} & 0.72 {\scriptsize[0.59,0.84]} & -- & 1.00 & 0.12\\
Claude Haiku 5.5 & 0.04 {\scriptsize[0.01,0.11]} & 0.06 {\scriptsize[0.02,0.13]} & 0.46 {\scriptsize[0.32,0.61]} & -- & 1.00 & 0.02\\
GPT-5.6 Sol$^\dagger$ & 0.12 {\scriptsize[0.05,0.21]} & 0.25 {\scriptsize[0.13,0.38]} & 0.86 {\scriptsize[0.75,0.94]} & -- & 0.97 & 0.18\\
Claude Opus 5$^\dagger$ & 0.09 {\scriptsize[0.03,0.17]} & 0.16 {\scriptsize[0.07,0.27]} & 0.72 {\scriptsize[0.56,0.86]} & -- & 0.85 & 0.06\\
\bottomrule
\end{tabular}
\end{table}

\paragraph{A new appearance costs almost nothing, a new level costs a lot.} A template keeps a
level's mechanics but can rotate or mirror the board, draw objects and the HUD in another style
and give roles other colours. Re-rendering one state under another template changes up to 99\% of
its cells (median over families of the largest change). Every episode carries
its renderer template, so success splits by appearance novelty and level novelty at once
(Table~\ref{tab:appearance}). Holding the levels and the start type fixed, a held-out template
costs written weights 1 point for Astra ($-0.01$ [$-0.03$, $+0.01$], paired over families), 2 for
Sol, 4 for Terra and 8 for Opus 5.5, and at most 5 in networks of the writers' own design. A
permuted palette, which changes the colour ids of every role, costs no more. The written actors
therefore locate objects from the frame rather than from fixed colours or positions, which they
could not read off the source (Section~\ref{sec:games}). Their first layers show how: 73\% of the
channels that read colour apply one kernel to eight or more colours or repeat one detector for each
of four or more colours, whereas every channel of a learned first layer reads all 16 colours with
distinct kernels. New levels are the difficulty: on the same
held-out templates, success falls from 78\% on support levels to 56\% on held-out ones for Astra,
and held-out levels between shown ones are solved more often than the final level (Astra W2 61\%
against 40\%, Sol 47\% against 31\%).

\paragraph{Ablations.} On six families with one run each (Table~\ref{tab:ablations},
Appendix~\ref{app:results}), solver code in the source does not clearly help written weights,
removing the actor's axial broadcasts changes little, and cutting its tied iterations from 128 to 8
hurts most.

\subsection{Weights versus programs (H3)}
\label{sec:h3}
\begin{figure}[t]
\centering
\begin{minipage}[c]{0.52\linewidth}
\includegraphics[width=\linewidth]{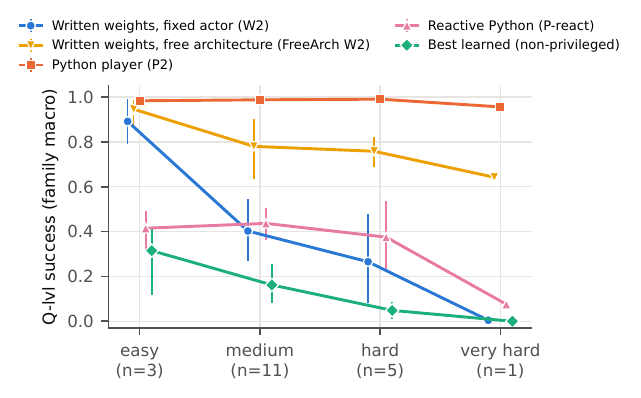}
\end{minipage}\hfill
\begin{minipage}[c]{0.44\linewidth}
\caption{Held-out-level success by difficulty stratum (family macro, bootstrap 95\% CI over
families, main writers pooled; the very hard stratum is one family). Written weights in the fixed
actor (W2) degrade with difficulty, reactive programs (P-react) stay near 40\% except on the very
hard family, free architectures (FreeArch W2) degrade less, and unrestricted programs (P2) not at
all. Learned baselines transfer worst.}
\label{fig:difficulty}
\end{minipage}
\end{figure}

Every writer's Python players beat its written weights: by 39--75 points one-shot, by 44--65 points
with feedback for the main writers, and in 76 of the 80 main writer--family pairs with
feedback (3 pairs tie and 1 is reversed). H3 was specified as a difficulty-by-format interaction
without a fixed test statistic, so we assess it descriptively per stratum
(Figure~\ref{fig:difficulty}; Table~\ref{tab:strata}, with CIs over families). The main writers'
written actors (W2) solve 89\% [79, 99] on easy families, 40\% [27, 55] on medium, 27\% [8, 48] on hard and
0\% on the very hard stratum, which is a single family. Python players stay at 96--99\% in every
stratum, so with P2 near ceiling the growing gap mainly reflects the decline of W2. An exploratory
trend test with the family as unit agrees: the per-family P2$-$W2 gap rises with stratum rank
(Spearman $\rho=0.67$, one-sided permutation $p<0.001$, 20 families; without the preliminary-study
families $\rho=0.59$, $p=0.007$). One-shot, where P1 is below ceiling
(Table~\ref{tab:strata}), it gives $\rho=0.52$ ($p=0.009$). H3 is supported descriptively.

P-react, the one-shot program control without simulation or search, solves 40\% pooled over the
main writers: close to written weights with feedback (42\%) and far below unrestricted programs
(99\%). It is not matched to the actor's computation (Appendix~\ref{app:results}).

What do these results say about writing weights? Across the nine writers and 20 families, a
one-shot network solves at least half of the held-out-level episodes only where the same writer's
one-shot Python player does too (19 of 19 such writer--family pairs in the fixed actor, 51 of 51 with
a free architecture, Section~\ref{sec:fa}). Where the program fails (34 pairs), the networks solve
2\% and 6\%. Understanding a game well enough to program it thus appears necessary for writing
its weights, and expressing that understanding in a fixed network is the hard part. This does not
show that writers derive their weights from a program, since any harder output format could give
the same pattern. Programs also handle new templates because the source contains the code of
the five support templates and the prompt says that evaluation uses unseen visual variants. Their players parse frames
invariantly, for example by taking the most frequent colour as background and matching objects by
shape rather than by colour id.

\subsection{A free architecture closes most of the gap (FreeArch, exploratory)}
\label{sec:fa}
Table~\ref{tab:fa_all} pairs each writer's FreeArch networks with its fixed-actor weights, and
Figure~\ref{fig:difficulty} shows both by difficulty. Every main writer gains 33--41 points with
feedback, all confidence intervals above zero, and FreeArch is better in 17--19 of the 20 families.
One-shot, FreeArch reaches
40--74\%. The collapse with difficulty largely disappears: pooled over the main writers, FreeArch
actors solve 95\% on easy, 78\% on medium, 76\% on hard and 64\% on the very hard stratum, where the
fixed actor reached 0\%. Without the
preliminary-study families, FreeArch W2 is 65--89\% and the gain 37--47 points, every interval above zero
(Table~\ref{tab:prelim}). As in the fixed actor, the final level is harder (Astra: 81\% against 93\%).

\begin{table}[t]
\centering\small\setlength{\tabcolsep}{4pt}
\caption{Free architecture (FreeArch) against the fixed actor on the 20 private families
(held-out-level success, family macro). $\Delta$ is the paired FreeArch$-$fixed difference with feedback
(W2), with a bootstrap 95\% CI over families; the last column counts families where FreeArch is better.
Upper block: main writers; lower block: weaker writers ($^\dagger$ previous generation).}
\label{tab:fa_all}
\begin{tabular}{lrrrrll}
\toprule
Writer & W1 fixed & W1 FreeArch & W2 fixed & W2 FreeArch & $\Delta$ W2 [95\% CI] & FreeArch$>$fixed\\
\midrule
GPT-6 Astra & 0.40 & 0.74 & 0.56 & 0.90 & +0.34 [+0.21,+0.48] & 17/20\\
GPT-6.1 Sol & 0.32 & 0.52 & 0.43 & 0.84 & +0.41 [+0.25,+0.56] & 17/20\\
Claude Opus 5.5 & 0.15 & 0.40 & 0.36 & 0.68 & +0.33 [+0.22,+0.45] & 18/20\\
GPT-6.1 Terra & 0.21 & 0.51 & 0.33 & 0.74 & +0.41 [+0.30,+0.52] & 19/20\\
\midrule
GPT-6 Luna & 0.05 & 0.09 & 0.11 & 0.14 & +0.04 [-0.01,+0.09] & 8/20\\
Claude Sonnet 5.5 & 0.06 & 0.17 & 0.19 & 0.39 & +0.20 [+0.11,+0.29] & 14/20\\
Claude Haiku 5.5 & 0.04 & 0.06 & 0.06 & 0.14 & +0.07 [+0.02,+0.13] & 12/20\\
GPT-5.6 Sol$^\dagger$ & 0.12 & 0.25 & 0.25 & 0.48 & +0.23 [+0.13,+0.32] & 17/20\\
Claude Opus 5$^\dagger$ & 0.09 & 0.21 & 0.16 & 0.33 & +0.17 [+0.07,+0.30] & 16/20\\
\bottomrule
\end{tabular}
\end{table}

\paragraph{Gap to programs.} Astra's FreeArch actors come within 10 points of its Python players, and
within 5 points on 12 of the 20 families. Sol 6.1 comes within 16 points, and Terra and Opus 5.5
within 24--28. Neither program control is matched to FreeArch: P2 may simulate and search without
bound, P-react may do neither, and FreeArch networks run bounded loops of up to 256 iterations.
With FreeArch's loop rules and caps, the bounded program players solve 91\%, against 79\% for FreeArch W2
($+0.11$ [$+0.06$, $+0.17$]; Table~\ref{tab:pbounded}), with positive point estimates for every
writer, although Sol 6.1's interval includes zero; unrestricted players stay 8 points ahead [4,
12]. The $+0.11$ estimates the cost of the weight format, with confounds in both directions
(Appendix~\ref{app:repro-actor}).

\paragraph{Where the gain comes from} (Table~\ref{tab:fa_controls}, Appendix~\ref{app:results}).
\begin{itemize}
\item \textbf{Not the prompt or the language.} Control A matches the fixed actor ($+0.01$ [$-0.02$, $+0.04$],
pooled over writers); all 240 delivered A graphs equal the fixed actor's graph.
\item \textbf{Mostly the topology.} Control B gains 29 points over A [21, 37], with every writer's
interval above zero. B is better in all 20 families pooled, and in 16, 19, 17 and 20 families for
Astra, Sol 6.1, Opus 5.5 and Terra. 11 of 608 B candidates used other operations and were rejected as invalid.
\item \textbf{The added operators contribute the rest} ($+0.06$ [$+0.02$, $+0.10$]), most clearly for
Opus 5.5 ($+0.12$).
\item \textbf{Not the parameter cap.} 152 of the 239 delivered FreeArch networks and 108 of the 240 B
networks stay within the fixed actor's 57k parameters and its MACs. These alone solve 80\% and 68\%
of held-out-level episodes, against 42\% for the fixed actor on the same families. This comparison
selects artifacts after the fact and is descriptive.
\end{itemize}
The controls do not separate the architecture from the weights written into it.

\paragraph{Weaker writers gain far less.} With feedback, the weaker writers' FreeArch networks solve
14--48\%, against 68--90\% for the main writers (Table~\ref{tab:fa_all}, lower block). FreeArch helps four of the five, by 7--23 points with confidence
intervals above zero, but not GPT-6 Luna ($+0.04$ [$-0.01$, $+0.09$]). The weaker writers can still write programs: one-shot, their
Python players solve 45--86\%, against 6--25\% for their FreeArch networks. Writing a working network from
the rules is thus concentrated in the strongest models.

\paragraph{What writers build.} The main writers' 451 delivered private-tier FreeArch networks
(one-shot and with feedback) have a median of 30k parameters (interquartile range 11k--64k) and
0.33~GMAC per step (0.07--0.80), less than the fixed actor. They rely heavily on multiplicative
gating, which the fixed actor lacks. 320 of them use tied loops of 2--256 iterations (median 48),
and 154 read the available-action mask. Their weights are hand-set constants: 93\% of all
parameters are exactly 0, and 90\% of the nonzero ones are $\pm1$, another small integer or a power
of two (97\% and 86\% for written fixed-actor weights, W2). They encode Boolean logic, for example NOT as $1-\min(\mathrm{relu}(x),1)$ and AND as a product. They are closed-loop: fed a constant frame instead of the observation, they solve at most 4\% of
held-out-level episodes. FreeArch
networks are thus bounded, branch-free tensor programs in network form, with fixed iteration counts.
This is close to compiling programs into networks~\citep{weiss2021thinking,lindner2023tracr},
except that the writer, not a compiler, derives the computation from informal rules and source.
Nor are they cheaper to run than programs: a FreeArch network takes a median of 5--24\,ms per step
on a GPU and 36\,ms on one CPU thread, against 0.8\,ms for a sandboxed Python player
(Table~\ref{tab:latency}, Appendix~\ref{app:results}).

\subsection{Written versus learned weights (H2)}
\label{sec:h2}
Table~\ref{tab:baselines} compares written weights with learned weights in the identical actor.
BC-support fits the support starts (95\%) but transfers poorly to held-out levels
(11\%). Distilling the writers' own Python players fits the teacher on the support levels
(92\% of support starts), yet solves only
10\% of held-out-level episodes, where the teachers solve over 99\% (Figure~\ref{fig:support},
Appendix~\ref{app:results}). PPO with sparse reward
reaches 8\%. The privileged PPO-shaped baseline fits the support set only partly and transfers no
better (5\%). With source-licensed augmentation, the learners fail to fit the support set within
our training budget: they solve only 17--20\% of support starts (BC-aug, distill-aug; 56\% with
dihedral transformations alone) and 7--10\% of held-out-level episodes. Per family, Astra's written
actors beat the best non-privileged learned baseline by 41 points [29, 54], and in 18 of the 20
families (both are near zero in the other two). H2 is supported, as a comparison of information access: writers read the source,
while learners see only the support levels (15--30 starts per family and their perturbations). The
learners with augmentation, which H2 names, did not fit their training levels, which weakens the
comparison. The result shows that support-level data do not convey what the source does; it is not a test of
learning as such.

The learned actors also fail without any new level. On the \emph{support} levels, where BC-support solves
92\% under the templates it trained on, a held-out template alone drops it to 13\%
($-0.79$ [$-0.86$, $-0.72$]; distillation $-0.74$, PPO $-0.34$; Table~\ref{tab:appearance}),
against $-0.01$ for Astra's written weights. The one learner that is appearance-robust,
BC-aug, buys it from palette and dihedral augmentation licensed by the source, and fits the support
set so poorly (20\%) that it gains nothing overall.

The actor can fit these episodes when trained on them. Trained with the planner teacher on states
from all levels, including the held-out levels' own start states (BC-oracle), it solves 90\%
of held-out-level episodes. This is an in-distribution capacity check, not generalisation, and it
isolates the appearance effect: BC-oracle trains on all ten templates and is robust to them
($+0.02$), yet a permuted palette, which it never sees, drops it from 90\% to 26\%, where written
weights lose nothing ($+0.01$ for Astra, paired over families). A learned actor of this architecture can therefore reach the written actors'
accuracy and still not survive a change of appearance.

\paragraph{Learned FreeArch networks.} A writer's own graphs learn almost nothing when trained from
standard initialisation. With re-initialised weights, Astra's FreeArch graphs solve 1\% of
held-out-level episodes under BC-support or distillation, against 89\% with Astra's written weights
($+0.89$ [$+0.77$, $+0.98$]; Table~\ref{tab:learned_fa}, Appendix~\ref{app:results}), although the
same learners fit the support set in the fixed actor. These graphs are hard-gated circuits built for
hand-set constants: many have no gradient at initialisation and others diverge. They are thus not
trainable from standard initialisation within our budget, which does not show that no learner could
fit them.

\begin{table}[t]
\centering\small
\caption{Learned baselines in the same 57k actor, family macro over the private families (seeds
averaged within a family, Appendix~\ref{app:repro-baselines}), against GPT-6 Astra W2. $^\ast$ privileged: PPO-shaped uses the
planner's distance as a potential, and BC-oracle trains on held-out levels.
The paired $\Delta$ compares Astra W2 with the best non-privileged baseline per family (bootstrap
over families).}
\label{tab:baselines}
\begin{tabular}{lrrrr}
\toprule
Method & S-start & Q-lvl & Q-tpl & families\\
\midrule
BC-support & 0.95 & 0.11 & 0.13 & 20\\
BC-aug & 0.20 & 0.09 & 0.21 & 20\\
BC-aug (dihedral only) & 0.56 & 0.10 & 0.25 & 13\\
Distill-aug & 0.17 & 0.07 & 0.17 & 20\\
Distill (no aug.) & 0.92 & 0.10 & 0.15 & 20\\
PPO & 0.47 & 0.08 & 0.12 & 20\\
PPO-shaped$^\ast$ & 0.33 & 0.05 & 0.07 & 6\\
BC-oracle$^\ast$ & 0.94 & 0.90 & 0.94 & 20\\
\midrule
GPT-6 Astra W2 & 0.80 & 0.56 & 0.78 & 20\\
\midrule
Paired $\Delta$ (W2 $-$ best non-privileged) & & 0.41 {\scriptsize[0.29,0.54]} & & 20\\
\bottomrule
\end{tabular}
\end{table}

\begin{table}[t]
\centering\small
\caption{Fragility of written weights: held-out-level success of the 18 written actors under noise
relative to each tensor's RMS, after PPO's first checkpoint and after a random update of the same
size (Tables~\ref{tab:perturbation} and~\ref{tab:fragility}).}
\label{tab:fragility_main}
\begin{tabular}{lrrrrr}
\toprule
 & \multicolumn{3}{c}{Relative noise} & PPO, first & Random update,\\
Written weights & 0 & $10^{-7}$ & $10^{-4}$ & checkpoint & same size\\
\midrule
Fixed actor, noise on all entries & 0.51 & 0.31 & 0.05 & 0.04 & 0.06\\
Fixed actor, noise on nonzero entries only & 0.51 & 0.45 & 0.08 & & \\
FreeArch networks, noise on all entries & 0.95 & 0.72 & 0.44 & & \\
\bottomrule
\end{tabular}
\end{table}

\subsection{Closed-loop behaviour and mechanism (H4)}
\label{sec:h4}
The main writers' W2 actors read the screen and act closed-loop (Table~\ref{tab:controls},
Appendix~\ref{app:results}). With a constant frame they solve at most 2\% of held-out
episodes, and frozen at the first frame 5--7\%. On perturbed starts without \texttt{RESET}
they solve 41--65\%, and their action changes on 71--76\% of counterfactual frames. H4 holds.

\paragraph{Mechanism.} In five strong written actors, we ablated the components the writers name,
after writing down the predicted effect. 14 of 17 predictions were confirmed: for example, zeroing
the tied-block channels a writer calls a flood fill drops held-out success from 0.97 to 0
(Appendix~\ref{app:results}).

\subsection{Public games}
\label{sec:public}
On the 25 public games, which are harder and possibly familiar (Tables~\ref{tab:public}
and~\ref{tab:fa_public}, Appendix~\ref{app:results}), fixed-actor written weights fail: at
most 8\% on held-out levels for every writer and condition. Free
architectures help (13--42\% with feedback, gains of 7--34 points), but they stay about 30 points
below the same writers' Python players (43--75\%). Possible familiarity does not rescue weight writing, and the larger residual gap
suggests that richer mechanics are harder to compile.

\subsection{Weight writing and standard PPO: no benefit in either direction}
\label{sec:rl}
\paragraph{From written weights to RL.} We fine-tuned the W2 actors of Astra, Sol 6.1 and Terra on
the six ablation families with PPO (3 seeds, 300k steps, sparse and planner-shaped reward;
Figure~\ref{fig:stage2}, Appendix~\ref{app:results}). Final actors are evaluated in exact FP32, and
diverged runs score 0. Written actors work on \texttt{ztl}, \texttt{lys} and \texttt{esm} (Astra's also on
\texttt{ruz}) and start near 0 on \texttt{crg} and \texttt{mds}, so the result rests mainly on three to
four families, three of them from the preliminary study. Standard PPO drives held-out success from 0.50 at initialisation to 0.01.
Structure-preserving PPO (Appendix~\ref{app:repro-baselines}) ends at
0.08--0.10, and small-step SGD at 0.22--0.25: success improves as the step size shrinks. RL from
the learned baselines' initialisation $\theta_0$ and from matched-random initialisations stays at
0.02 or below. Under the decision rule specified before data collection, no family, writer, variant
and reward combination beats both the written initialisation and RL($\theta_0$) (0 of 132).
Standard PPO collapses by its first checkpoint (16k steps), also in exact FP32 and on support
starts, and rescaling the output head does not help (Appendix~\ref{app:repro-baselines}). The main cause is fragility (Table~\ref{tab:fragility_main}): random noise of $10^{-7}$ of each tensor's
RMS already removes much of the success, at this size mostly through the exact zeros, although decisions are not
near ties. PPO's first checkpoint moves the weights by $10^{-4}$ overall, and a random update of the
same size and support is as destructive.

\paragraph{Robust written weights.} Asked for weights robust to noise of $10^{-3}$ of each tensor's
RMS, Astra and Sol 6.1 lose 3 and 8 points but keep most of their success up to that noise, about
$10^4$ times what halves standard weights (Table~\ref{tab:robust}; Opus 5.5 does not become
robust). PPO still removes their success (Table~\ref{tab:robust_ppo}): its first update on
them is 10--100 times larger, about as destructive as a random update of that size.

\paragraph{From RL to written weights.} Asked to repair RL-trained actors by editing weights,
Astra and Sol 6.1 rebuild them to the level of their fresh W2 actors, while continued PPO stays at
0.00--0.01 (Appendix~\ref{app:results}).

\section{Conclusions}
\label{sec:discussion}

Given the mechanics of an unpublished game, the strongest current language models can write the
weights of a network that plays new levels, renderings and start states of that game, closed-loop and far above replaying plans or repeating actions. Written weights also generalise better
than weights learned in the same network from the levels the writers saw, because the source tells
the writer what the support levels do not.

What the writers produce, however, are programs in network form. With a free architecture, the main
writers write bounded, branch-free tensor programs that come close to their own Python programs but
not beyond them. In the fixed architecture they fall far short: 145 of 240 main-writer feedback runs
stay below 50\% on held-out levels, although the same actor fits these episodes when trained on them.
Nor do written networks combine with training here: the architectures fail when trained from
scratch, and a single standard PPO update removes most of the fixed actor's success, even with
noise-robust weights. Where a task can be solved
by a program, as here, writing the program remains the better choice.

These results suggest a hypothesis about how weight-space programming works: the writer first
derives a symbolic solution and then expresses it in weights, so that what it writes is a program
even when the output is a network. Earlier demonstrations fit this reading.
\citet{takehi2026sparks} describe their writer as converting ``desired programs'' into neural form
for public games with known rules, and where no algorithm is known, written models are built from
hand-designed interpretable features~\citep{singh2026iwsp}. Their written transformers for
language tasks have the same signature as our
networks: rebuilding 21 of their 24 released language-task models from their code, we find 99\% of the
parameters exactly zero and 95\% of the rest small integers or powers of two
(Appendix~\ref{app:external-networks}). Our evidence comes from games whose
perception, although it must be computed at run time, can be specified exactly from the rendering
code, and whose policies admit a compact program. Tasks whose perception is statistical rather than
exact, such as natural images, would test the hypothesis, and tasks with dense rewards, where RL
fine-tuning works well, would test whether writing and training can combine.

\paragraph{Limitations.}
\begin{itemize}
\item \textbf{Scope.} We test generalisation within a game family, with the mechanics given as source code,
not the discovery of unknown rules or transfer to new games. Our games have binary success per
level, which makes them a poor testbed for combining writing with RL. The RL result covers
standard PPO on the fixed actor with from-scratch settings, with structure-preserving PPO and small-step SGD controls but without supervised
fine-tuning, and rests on three to four families (Section~\ref{sec:rl}).
\item \textbf{Estimation.} Per-family estimates rest on 3--6 held-out levels, and the ablations use one
run each. Learning budgets were bounded (2M PPO steps, 6 DAgger iterations, 6~h per learned FreeArch
network), so stronger learners might narrow the gap to written weights.
\item \textbf{Independence and elicitation.} The games were authored with language-model assistance
and never published. Prompts were piloted with evaluated writers, only the Claude writers get a
task-budget hint, and refused Opus 5.5 sessions score 0. An evaluated writer (Claude Opus 5.5)
designed the harness and analysis. Writers had no tools and had to
construct weights by design, which favours program-like networks.
\end{itemize}

\bibliographystyle{plainnat}
\bibliography{references}

\appendix
\section{Reproducibility details}
\label{app:repro}

This appendix specifies the actor, episodes, prompts, writer settings, sandbox, learned baselines,
evaluation sets and redaction procedure, and summarises evaluation numerics and integrity. All
evaluation sets and redacted game sections were fixed before the main study's first writer call, and
each condition's prompts before its first call.

\paragraph{Release.} Code and compact result data are available in the 
\href{https://github.com/JensLehmann/weight-space-programming-games}{project repository}
(commit \href{https://github.com/JensLehmann/weight-space-programming-games/tree/20eee0952edecc0b69865b84373d0345681f6e08}{\texttt{20eee09}}).
The release contains the evaluation and
writer harness, the code that builds every prompt, the free-architecture language with its
validator and executors, the public-tier query manifests, and per-session results and request
accounting for the main-study conditions, with a script that recomputes the headline numbers.
The robustness follow-up has aggregate noise curves, per-run PPO results and matched-random-update evaluations.
The sources and level data of the 20 private families are withheld so that they stay unseen. On
request, we evaluate submitted actors and players on them.

\subsection{Actor specification}
\label{app:repro-actor}

Table~\ref{tab:repro-actor} lists every tensor of the fixed actor. One step computes, with
$\star$ a convolution, $1\times1$ maps written as matrix products over channels, and
$T=128$:
\begin{align*}
F_1 &= \mathrm{ReLU}(W_1\star x_0+b_1), \qquad F_2=\mathrm{ReLU}(W_2\star F_1+b_2),\\
F_3 &= \mathrm{ReLU}\big(W_3\,[F_2;\,\mathrm{rowmax}(F_2);\,\mathrm{colmax}(F_2)]+b_3\big),\\
H^{(0)} &= 0,\qquad H^{(t)}=\mathrm{ReLU}\big(W_h\star H^{(t-1)}+U_hF_3+b_h\big),\quad t=1,\dots,T,\qquad H=H^{(T)},\\
g &= [\max H;\ \mathrm{mean}\,H;\ \max F_3;\ \mathrm{mean}\,F_3]\in\mathbb{R}^{96},\qquad z=\mathrm{ReLU}(W_zg+b_z),\\
m' &= \tanh(W_mm+W_{mz}z+W_{ma}a^-+b_m),\qquad \ell=W_o[z;m']+b_o,\qquad \kappa=W_c[F_3;H]+b_c.
\end{align*}
Here $\mathrm{rowmax}(F)_{c,i,j}=\max_{j'}F_{c,i,j'}$ and $\mathrm{colmax}(F)_{c,i,j}=\max_{i'}F_{c,i',j}$
broadcast each row's and column's maximum, $\max$ and $\mathrm{mean}$ in $g$ pool over the
$64\times64$ positions per channel (so $g$ has $16+16+32+32=96$ entries), $\ell$ are the eight action logits, and $\kappa$ is the
$64\times64$ click map. Convolution kernels are cross-correlations of shape
(out, in, $k$, $k$) with zero padding that keeps the $64\times64$ resolution. The input $x_0$ has
19 channels: 16 one-hot colour planes, row$/63$, column$/63$ and a constant 1. The previous-action
input $a^-$ is a one-hot over the eight actions (\texttt{RESET}, \texttt{ACTION1}--\texttt{ACTION7},
zero at episode start), and the memory $m$ is zero at episode start. Decoding is deterministic: the
argmax of $\ell$ over available actions (unavailable logits are $-\infty$, and ties go to the lowest
index), and, if \texttt{ACTION6} wins, the argmax cell of the click map $\kappa$ (ties go to the
first cell in row-major order). A step whose available-action logits are non-finite is recorded as
an invalid action rather than decoded.

\begin{table}[h]
\centering
\scriptsize
\caption{Tensors of the fixed actor (57{,}289 float32 parameters in total).}
\label{tab:repro-actor}
\begin{tabular}{llrl}
\toprule
Tensor & Shape & Entries & Role\\
\midrule
$W_1$, $b_1$ & $(32,19,5,5)$, $(32)$ & 15{,}232 & $5\times5$ convolution on $x_0$ $\to F_1$\\
$W_2$, $b_2$ & $(32,32,3,3)$, $(32)$ & 9{,}248 & $3\times3$ convolution $F_1\to F_2$\\
$W_3$, $b_3$ & $(32,96)$, $(32)$ & 3{,}104 & $1\times1$ mixing of $[F_2;\mathrm{rowmax}(F_2);\mathrm{colmax}(F_2)]\to F_3$\\
$W_h$ & $(16,16,3,3)$ & 2{,}304 & weight-tied $3\times3$ recurrence on $H$\\
$U_h$, $b_h$ & $(16,32)$, $(16)$ & 528 & $1\times1$ drive from $F_3$ into the recurrence\\
$W_z$, $b_z$ & $(128,96)$, $(128)$ & 12{,}416 & MLP on pooled features $g\to z$\\
$W_m$, $W_{mz}$, $W_{ma}$, $b_m$ & $(64,64)$, $(64,128)$, $(64,8)$, $(64)$ & 12{,}864 & episodic memory update (tanh)\\
$W_o$, $b_o$ & $(8,192)$, $(8)$ & 1{,}544 & action logits from $[z;m']$\\
$W_c$, $b_c$ & $(1,48)$, $(1)$ & 49 & click map from $[F_3;H]$\\
\midrule
Total & & 57{,}289 & \\
\bottomrule
\end{tabular}
\end{table}

\paragraph{Variants (operator ablation).} All variants share the tensor schema. \texttt{full}:
axial broadcasts and $T=128$ tied iterations. \texttt{noaxial}: $F_3=\mathrm{ReLU}(W_3[:,{:}32]\,F_2+b_3)$,
so the entries $W_3[:,32{:}]$ are ignored. \texttt{t8}: as \texttt{full} with $T=8$. Counting the
multiply-accumulates of convolutions and dense layers, a step costs 1.32, 1.31 and 0.19\,GMAC for
\texttt{full}, \texttt{noaxial} and \texttt{t8}. The FreeArch counting rule, which also counts
elementwise operations and reductions, gives 1.33 for \texttt{full}.

\paragraph{Numerics and initialisation.} Writers receive an exact NumPy reference of one actor step.
Evaluation uses an equivalent batched FP32 PyTorch implementation
(Appendix~\ref{app:integrity}). Tensors returned by a writer are cast to float32, and the number of
entries changed by the cast is recorded. Learned baselines start from $\theta_0$: weights drawn
uniformly from $[-1/\sqrt{f},\,1/\sqrt{f}]$, where $f$ is the fan-in of the tensor, zero biases,
and $W_h$ scaled by $0.1$ so that the 128-step recurrence starts stable.

\paragraph{Free-architecture language.} A FreeArch network is a list of nodes evaluated in order,
each applying one operation to the inputs or to earlier nodes, and a dictionary of named float32
arrays. The inputs are those of the fixed actor and the available-action mask. The operations are
convolutions with kernels of size 1, 3, 5 or 7, elementwise ReLU, tanh, sigmoid and absolute value,
the elementwise sum, difference, product, maximum and minimum of two tensors, channel concatenation
and slicing, row- and column-wise maximum and sum, global maximum and mean, dense layers, the
vector versions of the activations, sum, product, concatenation and slicing, and a loop that runs its body a fixed number of times (at most 256)
with tied weights. Loops cannot be nested, and the network contains no data-dependent control flow
and no argmax. A network may use at most 150k parameters, 1.5\,GMAC per step, 1{,}000 nodes,
10{,}000 node evaluations per step (a loop body counted per iteration), 512 channels per tensor,
4{,}096 channels in total and 256 memory units. A convolution or dense layer costs one MAC per
weight application, an elementwise binary operation one per output element and a reduction one per
input element. Outputs are decoded as for the fixed actor.

\paragraph{Bounded program player.} A NumPy function of the observation, the available-action
mask, a memory vector and the previous action returns action scores, click scores and the next
memory, decoded as for FreeArch. As in FreeArch, loops have fixed counts of at most 256 iterations and are not
nested, and there is no data-dependent Python control flow, scalar extraction or indexing. FreeArch's
caps apply to operations per call, element operations per call, array size ($512\times64\times64$
entries), memory and constants. Unlike FreeArch, the operation set also includes division, floor division, modulo,
powers, \texttt{exp}, comparisons, bitwise operators, \texttt{np.where}, \texttt{roll} and
\texttt{flip}, reshape and transpose, and reductions and contractions (with one constant operand)
along any axis, and FreeArch's limits of 1{,}000 nodes and 4{,}096 summed channels are not enforced. A
validator and a counting runtime enforce the rules, and violations are invalid artifacts. The
comparison with FreeArch is confounded in both directions: the bounded player has broader
operations, no node and channel limits and a familiar language, while the FreeArch prompt shows
the complete fixed actor and the bounded player's prompt only a trivial player.

\subsection{Episodes, actions and scoring}
\label{app:repro-episodes}

Observations are $64\times64$ grids of colour ids 0--15. A policy sees only the current frame and
the available-action mask. The game object and planner stay inside the evaluator. Level starts
are reached by continuous play from the first level, not by a debug initialiser. Every prompt
contains the following rules verbatim:
{\scriptsize
\begin{verbatim}
## Episodes and scoring

An episode starts at the beginning of one level (or at a state reached by a few actions into
it) and ends when that level is completed (success), on GAME_OVER, or after
3 x (shortest solution length) + 10 actions. Invalid actions count as wasted steps. RESET
restarts the current level and counts as an action. Success rate over many held-out episodes
is the score.
\end{verbatim}
}
\noindent The action cap of an episode is $3L+10$, where $L$ is the length of the planner's
certified direct plan from that episode's start state (Appendix~\ref{app:repro-sets}). Every
decision, including \texttt{RESET}, invalid actions, timeouts and errors, uses one step of the cap.
Losing a life ends an attempt but not the episode, and such events are counted. On the public tier the
engine is configured so that \texttt{RESET} restarts only the current level, as on the private tier
and as the rules state.

\subsection{Prompts}
\label{app:repro-prompts}

Prompts are identical across writers and differ between families only in the game section. In
order, a prompt contains a title, the game section (redacted simulator source and the engine base
class, Appendix~\ref{app:repro-redaction}), the episode rules above, the start observations of the
support set as hexadecimal grids (all support levels of template 0 and the first support level of
the other support templates) and the task.
\begin{itemize}
\item \textbf{Weights task.} The prompt gives the actor specification (prose, the NumPy reference
and the tensor shapes) and asks for \texttt{build()}, which returns every tensor with exactly its
shape. It states that the actor must play the game by itself on levels, layouts and visual
variants it has never seen, and that behaviour that only replays the support levels will fail. The
weights must be constructed by design: no training, optimising, fitting, searching over weights or
simulating the game inside \texttt{build()}. It receives no data, may use only \texttt{numpy},
\texttt{math} and pure-Python standard modules, and must finish within 60\,s. The reply must
explain the implemented mechanism (which channels and units compute what) and contain exactly one
Python code block.
\item \textbf{Python task.} It asks for \texttt{act(obs, available, memory)}, where \texttt{memory}
is a dictionary that persists within one episode, returning within 10\,s per step. P-react
additionally forbids reimplementing the transition function, simulating future states and searching
over action sequences, allows image analysis of the current frame, and sets 50\,ms per step.
\item \textbf{FreeArch and controls.} The FreeArch prompts replace the actor specification by the tensor
language, its caps and the fixed actor written in it. The prompts of controls A and B add one
paragraph with the graph or operation constraint, which the validator enforces. Operator-ablation
prompts add one paragraph that describes the variant.
\item \textbf{Repair of RL actors.} The task asks for \texttt{edit(params)} under the same
prohibitions and gives a per-tensor summary of the trained weights (shape, $\ell_2$ norm, maximum
magnitude, nonzero count). The full arrays are passed to \texttt{edit()} at run time. A second
interface (three runs per writer and family, like the first) adds the full values of the small
output, memory and readout tensors and per-channel statistics of the convolutional and encoder
tensors. Per family, Astra and Sol 6.1 reach 0.60--0.89 with the summary and 0.27--0.96 with the
values, and Opus 5.5 0.00--0.03 with either.
\end{itemize}

\paragraph{Feedback rounds (W2, FreeArch W2, P2).} After each submission the artifact is evaluated on
S-start (on the public tier, on S-start and S-dev). The feedback message contains a table
with template, level, start type, result and action count for every episode, and up to three
failure reports. Each report gives the reason (artifact error, invalid action, an action without
effect, a return to an earlier frame, or the cap), the first 40 actions, the frame at the first
problem, and, for weights, the logits, the value and location of the maximum of every $F_3$ and $H$
channel, and the memory norm at that frame. A session stops early once all support episodes are
solved. The final artifact is the candidate with the most support successes, ties going to the
latest round.

\subsection{Writer settings}
\label{app:repro-writers}

The writers are GPT-6 Astra, GPT-6.1 Sol, GPT-6.1 Terra, GPT-6 Luna and GPT-5.6 Sol (OpenAI) and
Claude Opus~5.5, Sonnet~5.5, Haiku~5.5 and Opus~5 (Anthropic), queried in November 2026 through
hosted APIs, with no tools, in stateless sessions: every turn resends the full visible transcript,
and no hidden reasoning is carried between turns.
\begin{itemize}
\item \textbf{Effort and output cap.} Reasoning effort \texttt{high} (\texttt{medium} in the
effort ablation) and an output cap of 128{,}000 tokens for every writer. A request is cancelled
after 90 minutes and not resubmitted.
\item \textbf{Claude writers} (Opus~5.5, Sonnet~5.5, Haiku~5.5, Opus~5): streaming, adaptive
thinking, and an advisory task budget of 96{,}000 tokens (beta \texttt{task-budgets-2026-03-13}).
The model sees a countdown of its remaining budget, which is not enforced. The enforced limit is the
128k output cap, of which the model is not informed.
\item \textbf{OpenAI writers} (GPT-6 Astra, GPT-6.1 Sol, GPT-6.1 Terra, GPT-6 Luna, GPT-5.6 Sol):
background requests, polled every 15\,s.
\item \textbf{Code extraction.} The last fenced Python block of the reply is executed, and a reply
without one is an invalid artifact.
\end{itemize}
Prompts were piloted on a toy game with all nine writers and on one family outside the study with
GPT-6 Astra, Claude Opus 5.5 and Claude Sonnet 5.5. A preliminary study then ran all nine writers
under W1 and P1, and the seven current ones also under W2 and P2, on the three study families named
in Section~\ref{sec:games} and another family outside the study, with the solver code in the source
and an earlier level split. Its results informed the protocol and prompts of the main study. No output from these runs entered any evaluated
prompt.

\subsection{Sandbox rules}
\label{app:repro-sandbox}

Writer code is audited statically and then executed in isolation. The threat model is a
non-adversarial writer that might accidentally read data, fit parameters or embed tables.
\begin{itemize}\sloppy
\item \textbf{Import allowlist} (checked in the AST and by an import hook at run time):
\texttt{numpy}, \texttt{math}, \texttt{itertools}, \texttt{functools}, \texttt{collections},
\texttt{heapq}, \texttt{random}, \texttt{dataclasses}, \texttt{typing}, \texttt{enum},
\texttt{copy}, \texttt{operator}, \texttt{statistics}, \texttt{bisect}, \texttt{time},
\texttt{array}, \texttt{re}, \texttt{string}, \texttt{fractions}, \texttt{numbers}, \texttt{abc},
\texttt{warnings}, \texttt{struct} and \texttt{decimal}. Relative imports are rejected.
\item \textbf{Rejected:} \texttt{open}, \texttt{exec}, \texttt{eval}, \texttt{compile},
\texttt{\_\_import\_\_}, \texttt{input}, \texttt{breakpoint}, \texttt{globals}, \texttt{locals},
\texttt{vars}, \texttt{memoryview}, \texttt{exit}, \texttt{quit}, \texttt{help}, \texttt{getattr},
\texttt{setattr}, \texttt{delattr}, \texttt{ctypes}, \texttt{system} and \texttt{popen}, every name
or attribute that starts with a double underscore, NumPy file input and output (\texttt{load},
\texttt{save}, \texttt{fromfile}, \texttt{tofile}, \texttt{memmap} and related) and a missing entry
function.
\item \textbf{Flagged (recorded, not rejected):} linear-algebra solvers and decompositions
(\texttt{lstsq}, \texttt{solve}, \texttt{pinv}, \texttt{inv}, \texttt{svd}, \texttt{eig} and
related), \texttt{while} loops, more than 1{,}000 numeric literals, and initialisers that use more
than 30\,s of CPU time. Flags are assigned by these static checks of the initialiser code, never
by hand. A flag never changes a score, and no artifact was excluded for one. Of the 3{,}719 delivered
network artifacts (final artifacts of fixed-actor, FreeArch and repair sessions), 210 carry a flag: 193 for
a \texttt{while} loop, 14 for numeric literals and 5 for a linear-algebra routine. Excluding them
changes no cell by more than 0.023.
\item \textbf{Isolation.} A separate \texttt{python -I} process in an empty working directory, a new
session, a scrubbed environment (single-threaded numerical libraries), restricted builtins, and no
data passed in: \texttt{build()} receives nothing and \texttt{edit()} only float32 copies of the
trained tensors.
\item \textbf{Time limits.} \texttt{build()} and \texttt{edit()}: 60\,s in the prompt, enforced as a
120\,s wall-clock and CPU limit. Python players run in one fresh process per episode, so no memory
survives between episodes, with 10\,s per step and 600\,s of CPU per episode. P-react: 50\,ms per step in
the prompt and a 100\,ms hard limit. P-react's other restrictions are stated in the prompt and
audited by source-pattern flags, not enforced by rejection: 56 of the 138 delivered P-react players
contain a flagged queue, heap or step function (the prompt allows distances and connected regions in
the current frame), and the main writers' flagged players score no higher than the others (0.38 against
0.44, mean over players).
A timeout, crash or exception inside \texttt{act} counts as an invalid action.
\item \textbf{Output validation.} The returned dictionary must contain exactly the tensor names of
Table~\ref{tab:repro-actor} (or of the writer's FreeArch graph), with exact shapes and finite values after
the float32 cast. Otherwise the artifact is invalid.
\end{itemize}

\subsection{Learned baselines}
\label{app:repro-baselines}

All fixed-architecture learned baselines train the actor of Table~\ref{tab:repro-actor} from $\theta_0$ and are
deployed with the same greedy decoding as written actors. Training episodes are S-start and S-dev
(support levels and templates only). BC-oracle instead uses every level and all ten templates,
each with its level start and three certified perturbed starts. Table~\ref{tab:repro-hparams}
lists the hyperparameters. Seeds per family are 3 for BC-support, BC-aug and PPO-shaped, 1 for
dihedral-only BC and BC-oracle, one student per Astra and Sol 6.1 run-0 player for both distillation
variants, and for PPO 3 on the six ablation families and 1 elsewhere (32 runs).

\paragraph{DAgger (BC and distillation).} DAgger~\citep{ross2011reduction} runs seven collection
rounds. Round 0 follows the teacher, and round $i\ge1$ follows it with probability $0.5^i$ and the
student otherwise. Every visited state is labelled by the teacher, and each round is followed by
four epochs of training. The loss is the cross-entropy of the teacher's action over available
actions plus, for \texttt{ACTION6} labels, the cross-entropy of the clicked cell over the click map.
States the planner teacher cannot label within its 5\,s budget are dropped from the loss, still feed
the memory, and are counted. The distillation teacher is the writer's final P2 player, run in the
sandbox.

\paragraph{Recurrent and progressive training.} Learners use truncated backpropagation through time
over windows of 16 steps through the episodic memory. Windows are cut from contiguous episode
segments with a random offset per episode and epoch, and are unrolled from the student's memory at the
window start. That memory is recomputed at every epoch by running the current student over the
same episode prefix, with the episode's augmentation applied to every frame, so a window never
starts from a memory computed on different frames or by an earlier student. A minibatch holds 4 windows for DAgger (64 frames) and
64 for PPO (1{,}024 frames). DAgger learners use a simplified
progressive loss~\citep{bansal2022end}: with probability 0.5 per minibatch, the tied block runs
$n\sim\mathcal{U}\{0,\dots,T-1\}$ iterations without gradient and $k\sim\mathcal{U}\{1,\dots,T-n\}$
with gradient, and the loss is computed on the output after $n+k$ iterations. Otherwise all $T=128$
iterations run with gradient. Deployment always uses $T=128$. PPO does not use the progressive loss,
because it would change the sampled policy and bias the importance ratio.

\paragraph{Source-licensed augmentation (BC-aug, distill-aug).} Two transformation families are
applied to training minibatches only, one per sample or window: (i) a random permutation of the 16
colour ids and (ii) one of the eight elements of the dihedral group $D_4$ acting on the frame, with
\texttt{ACTION1}--\texttt{ACTION4} remapped as screen directions, click coordinates mapped like
pixels, and the available-action mask and previous-action input permuted accordingly. A
transformation is licensed only when the game source shows it to preserve semantics. Palette
permutation is licensed for all private families. The dihedral group is licensed only where the
family's own templates apply board symmetries that generate $D_4$ and
\texttt{ACTION1}--\texttt{ACTION4} are absent or are screen directions: \texttt{lys}, \texttt{esm},
\texttt{aiv}, \texttt{obn}, \texttt{qwv}, \texttt{ubr}, \texttt{trx}, \texttt{qpv}, \texttt{yiy},
\texttt{crg}, \texttt{bvn}, \texttt{yuz} and \texttt{mds}.

\paragraph{PPO.} PPO~\citep{schulman2017proximal} with generalised advantage
estimation~\citep{schulman2015gae} and a separate critic used only in training: two $4\times4$
convolutions with stride 4 over the one-hot frame (32 and 64 channels), concatenated with the
actor's memory and previous action, and a 256-unit hidden layer. Episodes are drawn uniformly from
S-start and S-dev. Actions and click cells are sampled during rollouts, and evaluation is greedy. The
sparse reward is $+1$ on level completion. The shaped reward adds
$\gamma\Phi(s')-\Phi(s)$~\citep{ng1999policy} with $\Phi(s)=-d(s)$, the tightest certified planner
distance to go within 200\,ms, and $\Phi=0$ at every episode end. In all 18 PPO-shaped runs and all
shaped-reward runs from written initialisations, every potential was certified. Learning rate and
entropy coefficient were selected from $\{10^{-4},3\cdot10^{-4}\}\times\{0.003,0.01\}$ in 1M-step
runs on \texttt{lys}, \texttt{ruz} and \texttt{crg}, separately for each reward.

\begin{table}[h]
\centering
\scriptsize
\caption{Hyperparameters of the learned baselines.}
\label{tab:repro-hparams}
\begin{tabular}{lll}
\toprule
& Parameter & Value\\
\midrule
DAgger & collection rounds, mixing $\beta$ & 7 (round 0 teacher only), $0.5^i$\\
& epochs per round, optimiser & 4, Adam (fresh per round, lr $10^{-3}$)\\
& minibatch, gradient clip & 64 frames, norm 1.0\\
& buffer cap & 200{,}000 samples (whole random episodes dropped)\\
& planner teacher & 5\,s per state, no \texttt{RESET} fallback\\
& BPTT window, progressive loss & 16 steps, on ($p=0.5$)\\
& augmentation & licensed set (BC-aug, distill-aug), none otherwise\\
\midrule
PPO & environment steps & $2\times10^6$ (\texttt{bvn}, \texttt{qwv}: truncated at 1.80M, 1.06M)\\
& learning rate, entropy coefficient & $3\times10^{-4}$, 0.01\\
& parallel environments, horizon & 64, 128 (8{,}192 steps per update)\\
& epochs per update, minibatch & 4, 1{,}024 frames\\
& BPTT window & 16 steps\\
& $\gamma$, GAE $\lambda$, clip $\epsilon$ & 0.99, 0.95, 0.2\\
& value-loss coefficient, gradient clip & 0.5, norm 0.5 (actor and critic jointly)\\
& shaping (PPO-shaped) & planner potential, 200\,ms budget, scale 1.0\\
\bottomrule
\end{tabular}
\end{table}

\paragraph{RL from written initialisations.} On the six ablation families with three seeds, PPO
starts from the final W2 actor (run 0) of GPT-6 Astra, GPT-6.1 Sol or GPT-6.1 Terra, from
$\theta_0$, or from random weights matched to the written actor's per-tensor nonzero count and
$\ell_2$ norm, for 300k steps with the settings above and each reward. The variants are (a)
standard PPO, (b) structure-preserving PPO, which updates only entries that are nonzero at
initialisation, adds $0.05\,\mathrm{KL}(\pi\,\|\,\pi_{\text{init}})$ and trains only the critic for
the first 50k steps, (c) SGD on the actor with learning rate $10^{-5}$, and (d) standard PPO after
output-head calibration, which divides $W_o$ and $b_o$ so that the maximum absolute logit on support
starts is 5. Variant (d) runs for the 12 of 18 written actors whose divisor exceeds 1. Every
initialisation is evaluated under greedy and sampled decoding. Greedy checkpoints follow at the
first update boundary after 10k, 40k, 80k, 160k and 300k steps (16k, 41k, 82k, 164k and 303k). Of the 540 runs, 114 diverged to non-finite weights: 29 of 108
standard-PPO, 8 of 108 structure-preserving, 10 of 72 calibrated and 67 of 108 matched-random runs,
and none of the 108 SGD or 36 $\theta_0$ runs. Two properties of the written initialisations may
add to the collapse. Under the TF32 numerics of training, the written actors solve 56\% of support
starts greedily (70\% in exact FP32) but only 36\% when actions are sampled, the mode PPO trains
in, and 21\% of all runs diverge.

\subsection{Evaluation-set construction}
\label{app:repro-sets}

All sets are built mechanically from the games' own levels and renderer templates, without level
generators, and are fixed before the main study's first writer call. Table~\ref{tab:families} lists the private
families and Table~\ref{tab:repro-sets} the sets.
\begin{itemize}
\item \textbf{Templates and palettes} (private tier). Seed $s$ renders with template $s \bmod 10$.
Seeds 0--4 are support and seeds 5--9 held-out templates. Seeds 10--19 permute the palette and are
used only where the palette changes the frame. \texttt{trx} has eight templates, so only its seeds
5--7 count as held-out templates. The ten templates are a curated set, not a product of options:
in 16 of the 20 families at least one held-out template takes a value on some appearance axis that
no support template uses, and in 14 it uses a colour id that no support template uses. The colour
ids carrying a given role differ across templates in all 20 families and across palette seeds in
19. Re-rendering one state under another template changes 3--99\% of its cells (median over
families of the smallest and largest change).
\item \textbf{Perturbed starts.} From the level start, $k\sim\mathcal{U}\{1,\dots,6\}$ moves are
drawn uniformly from the moves that change the observation (clicks are probed on up to 48
non-background cells). A candidate is rejected if the prefix completes the level, leaves it or ends
an attempt, or if its frame duplicates an existing start. It is kept only if the planner's direct
plan from that state (no \texttt{RESET}, 5\,s budget) replays to completion in the native
environment. Each requested start gets at most 40 tries. Sets are drawn in the order S-dev, Q-off,
Q-lvl, Q-tpl with a shared exclusion set, so no two sets share a start state.
\item \textbf{Certification and cap.} Every start, including the level starts, must
certify. Its certified plan length $L$ sets the action cap $3L+10$. Planner outputs are used only to
certify starts and to supervise or shape learned baselines. No plan is shown to a writer.
\end{itemize}

\begin{table}[h]
\centering\small
\caption{The 20 private families: difficulty stratum (set by the game author before any result),
available actions besides RESET, held-out levels (1-based), number of levels, and evaluation-set
sizes (episodes). \mbox{S-start} has one start per support level and support template (5). Q-lvl has, per
held-out level and template (10), the level start and up to five perturbed starts. Q-tpl has, per
support level and held-out template (5), the level start and up to two perturbed starts. Sets fall
short of these counts where too few distinct perturbed starts could be drawn and certified within
40 tries each (Q-lvl of \texttt{ztl}, \texttt{vvf} and \texttt{yiy}, Q-tpl of \texttt{ztl}).}
\label{tab:families}
\begin{tabular}{llllrrrr}
\toprule
Family & Stratum & Actions & Held-out levels & Levels & S-start & Q-lvl & Q-tpl\\
\midrule
\texttt{ztl} & easy & A1,A2,A3,A4 & 2,4,6,8 & 8 & 20 & 210 & 50\\
\texttt{wih} & easy & A1,A5 & 2,4,6 & 6 & 15 & 180 & 45\\
\texttt{qcn} & easy & A1,A2,A3,A4,A5 & 2,4,6,8 & 8 & 20 & 240 & 60\\
\texttt{lys} & medium & A1,A2,A3,A4 & 2,4,6,8,10,12 & 12 & 30 & 360 & 90\\
\texttt{esm} & medium & A6 & 2,4,6,8,10 & 10 & 25 & 300 & 75\\
\texttt{aiv} & medium & A6 & 2,4,6,8,10,12 & 12 & 30 & 360 & 90\\
\texttt{nkp} & medium & A5,A6,A7 & 2,4,6,8,10 & 10 & 25 & 300 & 75\\
\texttt{obn} & medium & A1,A2,A3,A4,A5,A7 & 3,5,7,9 & 9 & 25 & 240 & 75\\
\texttt{qwv} & medium & A6,A7 & 2,4,6,8,10 & 10 & 25 & 300 & 75\\
\texttt{ruz} & medium & A1,A2,A6 & 2,4,6,8 & 8 & 20 & 240 & 60\\
\texttt{vvf} & medium & A1,A2,A3,A4,A5 & 2,4,6,8 & 8 & 20 & 209 & 60\\
\texttt{ubr} & medium & A6 & 3,5,7,9 & 9 & 25 & 240 & 75\\
\texttt{yuz} & medium & A5,A6,A7 & 2,4,6,8,10,12 & 12 & 30 & 360 & 90\\
\texttt{bvn} & medium & A6 & 3,5,7,9 & 9 & 25 & 240 & 75\\
\texttt{trx} & hard & A5,A6,A7 & 3,5,7,9,11 & 11 & 30 & 300 & 90\\
\texttt{yiy} & hard & A5,A6 & 2,4,6,8,10,12 & 12 & 30 & 341 & 90\\
\texttt{crg} & hard & A1,A2,A3,A4,A5 & 3,5,7,9 & 9 & 25 & 240 & 75\\
\texttt{qpv} & hard & A1,A2,A3,A4,A5 & 2,4,6,8 & 8 & 20 & 240 & 60\\
\texttt{ydv} & hard & A1,A2,A3,A4,A7 & 3,5,7,9,11 & 11 & 30 & 300 & 90\\
\texttt{mds} & very hard & A6,A7 & 3,5,7,9,11 & 11 & 30 & 300 & 90\\
\bottomrule
\end{tabular}
\end{table}

\begin{table}[h]
\centering
\scriptsize
\caption{Support and query sets per family.}
\label{tab:repro-sets}
\begin{tabular}{llll}
\toprule
Set & Levels & Templates / seeds & Starts per (level, template)\\
\midrule
\multicolumn{4}{l}{\emph{Private tier}}\\
S-start (feedback, training) & support & 0--4 & level start\\
S-dev (training only) & support & 0--4 & 4 perturbed\\
Q-off & support & 0--4 & 6 perturbed\\
Q-lvl (primary) & held out & 0--9 & level start + 5 perturbed\\
Q-tpl & support & 5--9 & level start + 2 perturbed\\
Q-pal & held out & palette seeds 10--19 & level start\\
\midrule
\multicolumn{4}{l}{\emph{Public tier (canonical rendering only)}}\\
S-start (feedback with S-dev, training) & support & canonical & level start\\
S-dev & support & canonical & 4 perturbed\\
Q-off & support & canonical & 6 perturbed\\
Q-lvl & held out & canonical & level start + up to 20 perturbed\\
\bottomrule
\end{tabular}
\end{table}

\paragraph{Public tier.} The 25 official games are provisioned at pinned versions. The longest
redacted section has about 126k tokens, mostly literal pixel arrays. Attempt ends are detected from the games' own signals, including lost lives. Every level
start certifies. Of the 3{,}353 perturbed-state certification checks recorded in the public
manifests, 2{,}675 succeed (79.8\%); duplicates and invalid perturbations are rejected before
certification. Only certified starts are evaluated. Q-lvl reaches 21 starts per held-out level
except for \texttt{g50t} level 3 (16 starts) and \texttt{wa30} level 9 (11 starts), where the
construction budget was exhausted. Augmentation licences come from a per-game source
survey and are conservative: palette permutation for 15 and the dihedral group for 4 of the 25
games.

\subsection{Redaction procedure}
\label{app:repro-redaction}

The prompt contains a redacted copy of each family's simulator source, while the harness keeps using the
full source to enter levels and certify starts. The redacted source keeps state, transitions, the
goal test, rendering, the level schema, support-level data and the engine base class. Each source
file is redacted in five steps:
\begin{enumerate}
\item Classes of other families that share the file are replaced by a one-line note.
\item Held-out entries of per-level data and renderer-template tables become \texttt{WITHHELD}
placeholders, which the prompt defines as evaluation-only entries. Branches conditioned on a
held-out level raise an error. Every function that matches solver, planner, oracle, search or
solution patterns is removed with every helper that only it uses, as are statements that store
solver output and comments that name solvers, optimal solutions or held-out levels. Mechanics gated
by a level index remain visible, because they are rules.
\item Solver-derived quantities in the mechanics are replaced. Oracle move budgets become explicit
per-level tables, with support values copied and held-out entries withheld. Support-level slack
fields are dropped, because budget minus slack would reveal the optimum. Loss rules or animations
that need a search call engine hooks, whose implementation is withheld and whose docstring states
exactly what they compute.
\item Comments and prose strings inside per-level data are removed, and the level data itself is unchanged.
\item Every top-level definition that the simulator class cannot reach is dropped (baselines,
diagnostics, probes, constants used only by held-out levels, unused imports).
\end{enumerate}
The M+S ablation applies the same redaction of levels and templates but keeps the solver code. The
game section opens with a fixed header that names the withheld levels and templates, states that
solver and planning code is not included, describes any engine hooks, and explains the start states
and template seeds.

\paragraph{Checks and audits.} Every redacted family must compile with no undefined names,
reproduce the original simulator frame for frame on every support level and template under
certified, perturbed and random action sequences, and raise an error for every held-out level and template. An
automated scan lists suspicious content (solver-like names, queue-based search over transition
functions, optimal move counts, conditionals on the level index, level numbers in comments,
per-level parameter tables, embedded solution witnesses). Two separate agent instances then reviewed
ten families each. These audits found leaks or broken code in 18 of
the 20 automatically redacted families, such as solvers retained because analysis code called them,
oracle-depth tables for held-out levels, held-out builder functions shipped in full and, in one
family, removed core mechanics. All were corrected before the main study's first writer call. Public-tier sections come
from reviewed per-game configurations plus a traced excerpt of the engine API and pass the same
equivalence check. A full read by separate agent instances found and fixed 10 further leaks. On the
public tier, generic rules used only by held-out levels and range gates that include a support level
are kept, so some held-out move budgets can be inferred.

\subsection{Evaluation and integrity}
\label{app:integrity}
\begin{figure}[!htbp]
\centering
\includegraphics[width=0.95\linewidth]{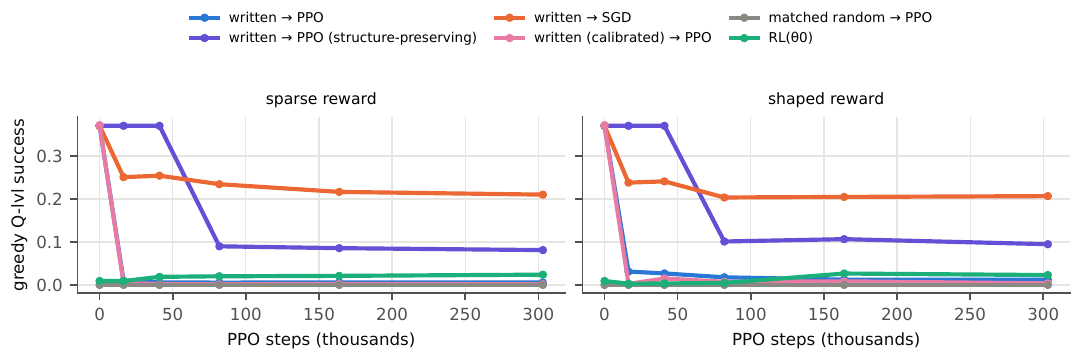}
\caption{Greedy held-out-level success during PPO from written initialisations (six families,
three writers, three seeds) and controls. Training-time checkpoints use TF32 convolutions, so the
step-0 checkpoint (about 0.37) is below the exact initial score (0.50). Diverged runs count as 0
after their last checkpoint.}
\label{fig:stage2}
\end{figure}

\begin{itemize}
\item \textbf{Exact FP32.} Final evaluation scores use exact FP32; Figure~\ref{fig:stage2} retains training-time TF32 checkpoint scores. With PyTorch's
default TF32 convolutions, 11 of 60 tested written actors change decisions, on 355 of 3{,}666
steps (9.7\%),
because written weights often rely on large gains and exact cancellations. The exact batched GPU
evaluator agrees with the NumPy reference on 99.7\% of 3{,}513 decisions from 60 written actors
(maximum relative logit error $8\times10^{-6}$, and the remaining differences are ties). Feedback during
writing used a backend that matches the reference within $3\times10^{-3}$. The mechanism ablations
also ran on it, because the batched CPU executor is not faithful (for one 128-step flood fill its logits
deviate by up to 10.6). Training of learned baselines and RL used TF32, and Section~\ref{sec:rl}
reports an exact-FP32 check.
\item \textbf{Re-runs.} A request was retried automatically only if it provably never started.
Whole sessions were re-run from scratch only after interruptions, never
based on their outcome: 310 of 5{,}900 sessions (329 discarded attempts), including 25 whose first
request failed at the API before any output. Excluding these sessions
changes no private-tier W2, P2 or FreeArch W2 cell by more than 0.006.
\item \textbf{Missing artifacts.} Refusals, truncated replies, invalid artifacts and undelivered
sessions score 0. In feedback sessions, a failed round keeps the best earlier artifact.
\item \textbf{Truncated runs.} Two PPO baseline runs were truncated, at 1.80M and 1.06M of
$2\times10^6$ steps (Table~\ref{tab:repro-hparams}).
\item \textbf{Accounting.} The main study used 11{,}637 writer requests in final sessions (588M
input, 209M output tokens). A usage-policy classifier stopped only Claude Opus 5.5, and only when it
wrote weights (17 of its 120 private-tier fixed-actor sessions, 9 other sessions and 3 of 42
RL-repair requests). Table~\ref{tab:requests} assigns every writer request of the main study
to a session. Of the 204 final sessions without an artifact, 168 returned an invalid artifact, 27
were refused, 8 ended with an API error and 1 was truncated.
Not counted are the 697 requests of the prompt pilots and the preliminary study, and 112 requests
of 28 sessions outside the design, which were discarded unevaluated. The robustness follow-up
in Appendix~\ref{app:robust} is reported separately and is not included in these request totals.
\end{itemize}

\begin{table}[t]
\centering\footnotesize\setlength{\tabcolsep}{3.5pt}
\caption{Main-study request accounting by arm. Wr.: writers. Req.: all writer requests, including discarded
attempts. Outcomes: completed, truncated (output cap), refused (usage policy), API error, and
interrupted (no outcome because the session was interrupted). Re-run: sessions
re-run from scratch after such an event. Only the final attempt is scored. $^a$Effort,
solver-source and operator ablations (W1/P1).}
\label{tab:requests}
\begin{tabular}{lrrrrrrrrrr}
\toprule
 &  &  &  & \multicolumn{5}{c}{Requests by outcome} & & No\\
\cmidrule(lr){5-9}
Arm & Wr. & Sess. & Req. & Compl. & Trunc. & Ref. & Err. & Interr. & Re-run & artifact\\
\midrule
\multicolumn{11}{l}{\emph{Private tier (20 families)}}\\
W1 & 9 & 540 & 571 & 530 & 0 & 9 & 1 & 31 & 31 & 40 (7\%)\\
P1 & 9 & 540 & 560 & 540 & 0 & 0 & 0 & 20 & 20 & 6 (1\%)\\
W2 & 9 & 540 & 2,065 & 2,015 & 1 & 8 & 4 & 37 & 38 & 11 (2\%)\\
P2 & 4 & 240 & 306 & 289 & 0 & 0 & 0 & 17 & 18 & 0 (0\%)\\
P-react & 7 & 140 & 141 & 140 & 0 & 0 & 0 & 1 & 1 & 2 (1\%)\\
Ablations$^a$ & 7 & 114 & 116 & 107 & 0 & 7 & 0 & 2 & 2 & 14 (12\%)\\
FreeArch W1 & 9 & 540 & 542 & 539 & 0 & 0 & 1 & 2 & 2 & 83 (15\%)\\
FreeArch W2 & 9 & 540 & 1,659 & 1,646 & 0 & 1 & 4 & 8 & 8 & 6 (1\%)\\
FreeArch control A & 4 & 240 & 840 & 840 & 0 & 0 & 0 & 0 & 0 & 0 (0\%)\\
FreeArch control B & 4 & 240 & 608 & 608 & 0 & 0 & 0 & 0 & 0 & 0 (0\%)\\
Bounded program & 4 & 240 & 443 & 415 & 0 & 0 & 22 & 6 & 21 & 1 (0\%)\\
RL-actor repair & 3 & 36 & 115 & 107 & 0 & 3 & 5 & 0 & 0 & 2 (6\%)\\
\midrule
\multicolumn{11}{l}{\emph{Public tier (25 games)}}\\
W1 & 7 & 525 & 558 & 526 & 0 & 0 & 0 & 32 & 32 & 21 (4\%)\\
P1 & 7 & 525 & 525 & 525 & 0 & 0 & 0 & 0 & 0 & 15 (3\%)\\
W2 & 4 & 300 & 1,386 & 1,338 & 0 & 1 & 4 & 43 & 96 & 1 (0\%)\\
P2 & 4 & 300 & 705 & 696 & 0 & 0 & 5 & 4 & 17 & 0 (0\%)\\
FreeArch W2 & 4 & 300 & 1,045 & 1,021 & 0 & 0 & 0 & 24 & 24 & 2 (1\%)\\
\midrule
Total & 9 & 5,900 & 12,185 & 11,882 & 1 & 29 & 46 & 227 & 310 & 204 (3\%)\\
\bottomrule
\end{tabular}
\end{table}

\renewcommand{\FloatBarrier}{}   \section{Additional results}
\label{app:results}

\begin{table}[!htbp]
\centering\footnotesize\setlength{\tabcolsep}{4pt}
\caption{Appearance novelty against level novelty (private tier, family macro, with feedback).
Every episode carries its renderer template: support templates are the five the writer sees
through the simulator source, new templates are the five withheld from it, and a permuted palette
re-draws the colour ids of a template the writer does see. ``$\Delta$ appearance'' is new minus
support templates on the support levels, paired over families with a bootstrap 95\% CI. It holds
the levels fixed, and separating level starts from perturbed starts changes it by at most 0.01. The
palette column holds level starts only, on the 19 families whose palette changes the frame.
Learned baselines train on support levels under support templates. $^\ast$BC-oracle trains on
held-out levels under all ten templates, so only its palette column is out of distribution.}
\label{tab:appearance}
\begin{tabular}{lrrlrrr}
\toprule
& \multicolumn{3}{c}{Support levels} & \multicolumn{3}{c}{Held-out levels}\\
\cmidrule(lr){2-4}\cmidrule(lr){5-7}
Policy source & Support & New & $\Delta$ appearance [95\% CI] & Support & New & Permuted\\
& templ. & templ. & & templ. & templ. & palette\\
\midrule
GPT-6 Astra W2 & 0.79 & 0.78 & -0.01 {\scriptsize[-0.03,+0.01]} & 0.57 & 0.56 & 0.54\\
GPT-6.1 Sol W2 & 0.62 & 0.60 & -0.02 {\scriptsize[-0.04,0.00]} & 0.44 & 0.43 & 0.39\\
Claude Opus 5.5 W2 & 0.51 & 0.43 & -0.08 {\scriptsize[-0.13,-0.03]} & 0.39 & 0.33 & 0.31\\
GPT-6.1 Terra W2 & 0.52 & 0.48 & -0.04 {\scriptsize[-0.09,-0.01]} & 0.35 & 0.32 & 0.30\\
\midrule
GPT-6 Astra FreeArch W2 & 0.97 & 0.97 & 0.00 {\scriptsize[-0.01,0.00]} & 0.90 & 0.90 & 0.90\\
GPT-6 Astra P2 (Python) & 1.00 & 1.00 & 0.00 {\scriptsize[0.00,0.00]} & 1.00 & 1.00 & 1.00\\
\midrule
BC-support & 0.92 & 0.13 & -0.79 {\scriptsize[-0.86,-0.72]} & 0.15 & 0.07 & 0.07\\
Distill (no aug.) & 0.89 & 0.15 & -0.74 {\scriptsize[-0.82,-0.65]} & 0.14 & 0.07 & 0.08\\
PPO & 0.46 & 0.12 & -0.34 {\scriptsize[-0.45,-0.24]} & 0.11 & 0.05 & 0.06\\
BC-aug & 0.21 & 0.21 & 0.00 {\scriptsize[-0.02,+0.01]} & 0.09 & 0.09 & 0.08\\
BC-oracle$^\ast$ & 0.92 & 0.94 & +0.02 {\scriptsize[0.00,+0.04]} & 0.89 & 0.90 & 0.26\\
\bottomrule
\end{tabular}
\end{table}

\begin{table}[t]
\centering\footnotesize\setlength{\tabcolsep}{3pt}
\caption{Fragility of written fixed-actor weights (W2 run 0 of Astra, Sol 6.1 and Terra on the six
ablation families, 18 actors). Isotropic Gaussian noise of $\epsilon$ times each tensor's RMS, RL
checkpoints, and random updates with the same size and support as each RL update. The checkpoints
are those of the seed-0 sparse-reward runs of Section~\ref{sec:rl}, reproduced in exact FP32, and
update sizes are medians over actors. Held-out success uses 60 episodes per actor (the same
subsample throughout, so values differ slightly from Section~\ref{sec:rl}) and support starts are
evaluated in full, with family bootstrap 95\% CIs.}
\label{tab:perturbation}
\begin{tabular}{lrrll}
\toprule
Weights ($n$ actors) & \multicolumn{2}{c}{$\|\Delta\theta\|/\|\theta\|$} & Q-lvl [95\% CI] & S-start [95\% CI]\\
 & overall & max.\ tensor & & \\
\midrule
Written W2 actor (18) & 0 & 0 & 0.51 [0.18, 0.82] & 0.70 [0.50, 0.89]\\
\quad + isotropic noise, $\epsilon=10^{-7}$ & $10^{-7}$ & $1.2\cdot10^{-7}$ & 0.31 [0.11, 0.55] & 0.34 [0.11, 0.63]\\
\quad + isotropic noise, $\epsilon=10^{-6}$ & $10^{-6}$ & $1.2\cdot10^{-6}$ & 0.24 [0.08, 0.46] & 0.27 [0.07, 0.55]\\
\quad + isotropic noise, $\epsilon=10^{-5}$ & $10^{-5}$ & $1.2\cdot10^{-5}$ & 0.10 [0.02, 0.20] & 0.17 [0.05, 0.33]\\
\quad + isotropic noise, $\epsilon=10^{-4}$ & $10^{-4}$ & $1.2\cdot10^{-4}$ & 0.05 [0.00, 0.13] & 0.12 [0.02, 0.26]\\
\quad + isotropic noise, $\epsilon=10^{-3}$ & $10^{-3}$ & $1.2\cdot10^{-3}$ & 0.04 [0.00, 0.10] & 0.10 [0.01, 0.23]\\
\quad + isotropic noise, $\epsilon=10^{-2}$ & $10^{-2}$ & $1.2\cdot10^{-2}$ & 0.02 [0.00, 0.05] & 0.04 [0.01, 0.09]\\
\quad + isotropic noise, $\epsilon=10^{-1}$ & $10^{-1}$ & $1.2\cdot10^{-1}$ & 0.01 [0.00, 0.03] & 0.03 [0.01, 0.07]\\
\midrule
Standard PPO, first ckpt 16k (18) & $9.4\cdot10^{-5}$ & $2.0\cdot10^{-2}$ & 0.04 [0.00, 0.12] & 0.08 [0.00, 0.19]\\
\quad random direction, same size (15) & & & 0.06 [0.00, 0.14] & 0.07 [0.00, 0.18]\\
\midrule
Structure-preserving PPO, first ckpt after warm-up 82k (3) & $2.0\cdot10^{-5}$ & $2.5\cdot10^{-2}$ & 0.26 [0.00, 0.52] & 0.42 [0.25, 0.57]\\
\quad random direction, same size (3) & & & 0.11 [0.00, 0.22] & 0.18 [0.02, 0.40]\\
\quad same actors at initialisation & & & 0.91 [0.82, 1.00] & 1.00 [1.00, 1.00]\\
\midrule
SGD, lr $10^{-5}$, final 303k (18) & $4.2\cdot10^{-8}$ & $1.5\cdot10^{-5}$ & 0.29 [0.10, 0.46] & 0.33 [0.13, 0.51]\\
\quad random direction, same size (18) & & & 0.27 [0.10, 0.44] & 0.31 [0.12, 0.50]\\
\midrule
Structure-preserving PPO, final 303k (17) & $1.4\cdot10^{-5}$ & $4.2\cdot10^{-2}$ & 0.10 [0.03, 0.17] & 0.17 [0.07, 0.29]\\
\quad random direction, same size (17) & & & 0.07 [0.00, 0.16] & 0.11 [0.02, 0.25]\\
\quad same actors at initialisation & & & 0.50 [0.18, 0.81] & 0.69 [0.50, 0.89]\\
\bottomrule
\end{tabular}
\end{table}

\begin{table}[t]
\centering\footnotesize\setlength{\tabcolsep}{3pt}
\caption{Noise on all entries and on nonzero entries only (same overall relative size $\epsilon$), for
the fixed actors of Table~\ref{tab:perturbation} and the final FreeArch W2 networks (run 0) of the same
writers and families (18 each). Held-out success uses the same subsample. The first row is from
Table~\ref{tab:perturbation} (two noise directions), and the other rows use one direction. Without noise, exact
ties between the two largest action logits occur in 1.6\% (fixed actor) and 12\% (FreeArch) of decisions,
the median gap between them is 0.99 and 0.94 of the larger one, and reversing the tie-break changes
62 and 24 of 5{,}400 episode outcomes. FreeArch networks differ widely: 5 of 18 lose more than half their success at $10^{-7}$, 5 keep
their success up to $10^{-4}$.}
\label{tab:fragility}
\begin{tabular}{lrrrrrrrr}
\toprule
Held-out Q-lvl at relative noise $\epsilon$ & 0 & $10^{-7}$ & $10^{-6}$ & $10^{-5}$ & $10^{-4}$ & $10^{-3}$ & $10^{-2}$ & $10^{-1}$\\
\midrule
Fixed actor, all entries & 0.51 & 0.31 & 0.24 & 0.10 & 0.05 & 0.04 & 0.02 & 0.01\\
Fixed actor, nonzero entries only & 0.51 & 0.45 & 0.31 & 0.20 & 0.08 & 0.06 & 0.05 & 0.02\\
FreeArch network, all entries & 0.95 & 0.72 & 0.74 & 0.62 & 0.44 & 0.24 & 0.03 & 0.02\\
FreeArch network, nonzero entries only & 0.95 & 0.71 & 0.72 & 0.62 & 0.54 & 0.24 & 0.05 & 0.02\\
\bottomrule
\end{tabular}
\end{table}

\paragraph{External language-task networks.}\label{app:external-networks}
For the comparison with \citet{singh2026iwsp}, we inspected their repository snapshot retrieved on
10 November 2026. The sample comprises \texttt{sentiment-sst2}, \texttt{nli-snli},
\texttt{paraphrase-mrpc} and \texttt{trec-coarse}, each with \texttt{haiku55}, \texttt{opus55}
and \texttt{sonnet55} runs \texttt{s1} and \texttt{s2} (24 models). We called each model's
\texttt{build\_model()}, disabled the separate \texttt{dev\_report} diagnostic, and ran no task
evaluation. Three constructors failed because the \texttt{datasets} dependency was unavailable:
\texttt{nli-snli-opus55-s2}, \texttt{nli-snli-sonnet55-s1} and \texttt{trec-coarse-sonnet55-s2}.
The 21 constructed models contain 252{,}513{,}103 parameters. Shares are pooled over parameters,
with exact zero, $\pm1$, other nonzero integers of magnitude at most 16, and powers of two of
magnitude $2^{-8}$--$2^{16}$ classified as in our own weight analysis. The pooled zero share is
about 99.2\%, and about 95.4\% of the nonzero parameters fall in the latter three categories.

\paragraph{Robust written weights.}\label{app:robust} This condition is the W2 session of the main
study (fixed actor, mechanics-only source, three feedback rounds) with one added rule: the actor
must keep playing well when every tensor receives independent Gaussian noise with standard
deviation $10^{-3}$ times its RMS, added to every entry including the zeros. The prompt adds that
weights relying on exact cancellations, exact zeros or very large gains usually fail under such
noise. Feedback reports the support success of the exact actor and of the actor under one noise
draw, with a new draw in every round, and the final round is the one with the most exact plus noisy
support successes. Astra, Sol 6.1 and Opus 5.5 wrote three runs on each of the six ablation families.
The API repeatedly returned no response for Sol 6.1 on \texttt{crg} and \texttt{ruz}, so Sol 6.1 has one run
on \texttt{crg} and none on \texttt{ruz}. Opus 5.5 sessions stopped by the usage-policy classifier (4 of
18) score 0. Table~\ref{tab:robust} compares the noise curves with the standard W2 runs of the same
writers, families and run indices, and Table~\ref{tab:robust_ppo} gives standard PPO from run 0 of
both prompts (exact FP32, 300k steps, sparse and planner-shaped reward, three seeds).

PPO's first update moves the robust weights by $1.0\cdot10^{-3}$ of their norm (median over the 15
run-0 actors with both prompts), against $2.9\cdot10^{-5}$ for the standard ones (median over the 12 whose first update stays
finite, while the other three diverge to non-finite weights), with
similar largest per-tensor changes ($1.2\cdot10^{-2}$ and $1.1\cdot10^{-2}$). On the same actors, held-out
success is 0.48 at initialisation (standard 0.57), 0.45 under noise of $10^{-4}$ (0.05), 0.10 after
PPO's first update (0.04, with the three diverged runs scored 0) and 0.14 after a random update with
the same per-tensor size and support (0.05, over the 12 finite updates). Robust actors are much weaker when actions are sampled, the mode PPO trains
in (Astra 0.62 greedy and 0.15 sampled, against 0.61 and 0.40 for its standard actors), but rescaling
the output head to a largest logit of 5 does not help: on the nine robust actors that work, held-out
success falls from 0.92 to 0.07 after the first update. Robust runs diverge to non-finite weights
less often (6 of 96 against 23 of 102), and none beats both its initialisation and RL($\theta_0$).

\begin{table}[t]
\centering\footnotesize\setlength{\tabcolsep}{3pt}
\caption{Noise curves of robust and standard written fixed-actor weights (held-out-level success,
family macro over the ablation families, matched run indices). The noise-free column uses the full
held-out set, the others the fixed subsample of Table~\ref{tab:perturbation}, with two noise draws.
Refused Opus 5.5 sessions score 0.}
\label{tab:robust}
\begin{tabular}{llrrrrrrrr}
\toprule
Prompt & Writer (families) & 0 & $10^{-7}$ & $10^{-6}$ & $10^{-5}$ & $10^{-4}$ & $10^{-3}$ & $10^{-2}$ & $10^{-1}$\\
\midrule
Standard & GPT-6 Astra (6) & 0.63 & 0.35 & 0.24 & 0.10 & 0.05 & 0.05 & 0.03 & 0.01\\
Robust & GPT-6 Astra (6) & 0.60 & 0.59 & 0.59 & 0.59 & 0.59 & 0.53 & 0.10 & 0.01\\
\midrule
Standard & GPT-6.1 Sol (5) & 0.56 & 0.26 & 0.14 & 0.09 & 0.03 & 0.03 & 0.01 & 0.01\\
Robust & GPT-6.1 Sol (5) & 0.48 & 0.48 & 0.48 & 0.48 & 0.48 & 0.41 & 0.04 & 0.01\\
\midrule
Standard & Claude Opus 5.5 (6) & 0.41 & 0.19 & 0.15 & 0.12 & 0.03 & 0.03 & 0.00 & 0.01\\
Robust & Claude Opus 5.5 (6) & 0.23 & 0.23 & 0.21 & 0.21 & 0.18 & 0.12 & 0.05 & 0.02\\
\bottomrule
\end{tabular}
\end{table}

\begin{table}[t]
\centering\footnotesize\setlength{\tabcolsep}{4pt}
\caption{Standard PPO from robust and standard written weights (run 0, exact FP32): held-out-level
success by training step, mean over seeds and rewards, family macro over the families where both
prompts produced an actor. Diverged runs score 0 from their divergence on.}
\label{tab:robust_ppo}
\begin{tabular}{llrrrrrrr}
\toprule
Prompt & Writer (families) & 0 & 8k & 16k & 41k & 82k & 164k & 303k\\
\midrule
Standard & GPT-6 Astra (6) & 0.61 & 0.02 & 0.01 & 0.01 & 0.01 & 0.01 & 0.01\\
Robust & GPT-6 Astra (6) & 0.62 & 0.07 & 0.03 & 0.00 & 0.00 & 0.01 & 0.00\\
\midrule
Standard & GPT-6.1 Sol (5) & 0.59 & 0.01 & 0.01 & 0.01 & 0.01 & 0.01 & 0.01\\
Robust & GPT-6.1 Sol (5) & 0.44 & 0.05 & 0.05 & 0.04 & 0.03 & 0.03 & 0.03\\
\midrule
Standard & Claude Opus 5.5 (4) & 0.48 & 0.03 & 0.03 & 0.03 & 0.03 & 0.03 & 0.02\\
Robust & Claude Opus 5.5 (4) & 0.33 & 0.09 & 0.09 & 0.08 & 0.09 & 0.10 & 0.10\\
\bottomrule
\end{tabular}
\end{table}

\begin{table}[t]
\centering\small\setlength{\tabcolsep}{4pt}
\caption{Bounded program player against FreeArch and the fixed actor (held-out-level success with
feedback, family macro over the 20 private families). Bounded: NumPy players with FreeArch's loop rules
and caps and a broader operation set (Appendix~\ref{app:repro-actor}). Differences are paired over families with bootstrap 95\% CIs. The pooled row
averages the four writers per family first.}
\label{tab:pbounded}
\begin{tabular}{lrrrrll}
\toprule
Writer & Fixed W2 & FreeArch W2 & Bounded & P2 & Bounded$-$FreeArch & P2$-$Bounded\\
\midrule
GPT-6 Astra & 0.56 & 0.90 & 0.98 & 1.00 & +0.08 [+0.02,+0.15] & +0.02 [+0.00,+0.04]\\
GPT-6.1 Sol & 0.43 & 0.84 & 0.91 & 1.00 & +0.07 [-0.02,+0.16] & +0.09 [+0.03,+0.17]\\
Claude Opus 5.5 & 0.36 & 0.68 & 0.87 & 0.97 & +0.18 [+0.11,+0.26] & +0.10 [+0.06,+0.14]\\
GPT-6.1 Terra & 0.33 & 0.74 & 0.87 & 0.98 & +0.13 [+0.02,+0.23] & +0.11 [+0.04,+0.21]\\
\midrule
Pooled (4 writers) & 0.42 & 0.79 & 0.91 & 0.99 & +0.11 [+0.06,+0.17] & +0.08 [+0.04,+0.12]\\
\bottomrule
\end{tabular}
\end{table}

\paragraph{Ablations.} In Table~\ref{tab:ablations}, solver code in the source gives Astra 0.53,
Sol 0.41 and Opus 0.21, each within the range of single runs without it (Astra 0.22--0.56, Sol
0.31--0.43, Opus 0.06--0.30). Without axial broadcasts, Astra still solves 0.50 and Sol 0.33. With 8
instead of 128 tied iterations, Astra drops to 0.16 and Sol to 0.28, while Opus 5.5 does not drop
(0.23, with 3 of 6 sessions delivered). With effort \texttt{medium}, Astra and Sol lose 4--10
points and Terra drops from 0.27 to 0.09. Opus 5.5 scores 0.18 against 0.15, and the three other
current writers stay below 0.08 at either effort.

\paragraph{Learned FreeArch networks.} In Table~\ref{tab:learned_fa}, Astra's FreeArch graphs with
re-initialised weights solve 1\% of support starts and 1\% of held-out-level episodes under either
learner, against 100\% and 89\% with the written weights (higher in 19 of 20 families, tied in the
family where the written network also fails). The same learners fit the support set in the fixed
actor (90--96\%). With random weights, 9 of the 20 graphs have at least 99\% of weight entries
without gradient on support frames, because an early clamp is inactive everywhere, and one is
numerically unstable from the start. Four BC and five distillation runs diverged and score 0, and
seven reached the time limit.

\paragraph{Cost and latency.}\label{sec:cost} We timed 120 FreeArch actors drawn at random from the 451 above, 60
fixed-actor written actors and 60 Python players, in exact FP32 on one otherwise idle GPU
(Table~\ref{tab:latency}, Appendix~\ref{app:results}). The FreeArch networks have fewer parameters and
MACs than the fixed actor, but they evaluate a median of 2{,}362 small nodes per step, against 412.
At batch 1 in matched eager mode they are 5$\times$ slower (24 against 4.8\,ms); batched, they cost
525\,$\mu$s per action against 273\,$\mu$s. Sandboxed Python players take a median of 0.8\,ms per
call on one CPU thread, where both network types need about 36\,ms. Written networks are thus not
cheaper per step than programs, only static and bounded.

\paragraph{Reactive programs (P-react).} P-react is a one-shot program control without simulation or search (one run per family;
Appendix~\ref{app:repro-sandbox}). Pooled over
the main writers, it solves 40\%: above one-shot written weights (27\%), close to written
weights with feedback (42\%), and far below unrestricted programs (99\%). Because P-react forbids
search while the fixed actor iterates its tied block 128 times, it is not matched to the actor's
computation and does not separate understanding the rules from expressing them.

\paragraph{Structure of the Python players.} Of the final unrestricted programs (P2), only 7--39\%
(by writer) contain explicit queue-based search and up to 29\% re-implement transition steps. Most
parse the state and compute the answer directly.

\paragraph{Mechanism.} In five strong written actors (Astra and Sol 6.1; two easy, two medium and
one hard family), we ablated the components the writers name, after writing down the predicted
effect. 14 of 17 predictions were confirmed; each of the 3 misses failed on one metric. Zeroing the tied-block channels the writer calls a flood fill, in a maze family, drops held-out
success from 0.97 to 0. In the hard family, zeroing the channels described as value propagation
drops it from 1.0 to 0.04. Zeroing one direction latch in memory leaves unperturbed starts at
1.0 but halves success on perturbed starts. The networks are very sparse: 1.4--13.8\% of weights are nonzero, the tied block uses 2--16 channels and memory at most 7 of 64 units. FreeArch networks are closed-loop as well (family means per writer): frozen at the first frame, they solve 5--8\% of held-out-level episodes, and the chosen action changes on 72--76\% of counterfactual frames.

\paragraph{From RL to written weights.} Asked to repair the 2M-step PPO-shaped actors of two families
(held-out success 0.00 and 0.03) through \texttt{edit(params)}, shown a per-tensor summary or also
the small tensors' values (Appendix~\ref{app:repro-prompts}), Astra and Sol 6.1 rebuild them
(21 of 23 actors change more than 99.9\% of entries) and reach 0.27--0.96, like their
fresh W2 actors (0.32--0.92); Opus 5.5 makes small edits and never exceeds the RL actor. Continued
PPO stays at 0.00--0.01.

\paragraph{Preliminary-study families.} Table~\ref{tab:prelim} repeats the main comparisons
without the three study families of the preliminary study. On the remaining 17 families, the H3
trend test gives Spearman $\rho=0.59$ (one-sided permutation $p=0.007$), against $\rho=0.67$
($p<0.001$) on all 20. The three families are 3 of the 6 ablation families and 3 of the 5 mechanism
actors (Section~\ref{sec:h4}). On the other three ablation families (\texttt{ruz}, \texttt{crg},
\texttt{mds}), one-shot written weights solve at most 0.13 in every ablation arm, so the ablation
contrasts come almost entirely from the preliminary-study families.

\paragraph{Protocol-specified secondary analyses.} Table~\ref{tab:prespec} reports delivery, the
analyses conditional on delivery and without undelivered sessions, Q-pal and the number of runs
that reach 0.5. With feedback, only Opus 5.5 has undelivered sessions (8 in W2, 1 in FreeArch W2).
Without them, its W2 rises from 0.36 to 0.39--0.41. In the one-shot arms, where delivery is
lower, excluding undelivered sessions raises Opus 5.5's W1 from 0.15 to 0.18 and Sol 6.1's FreeArch W1
from 0.52 to 0.67, and changes every other main-writer cell by at most 0.03. Q-pal is within 5 points
of Q-lvl in every cell, but it contains only level starts and 19 families, so the difference is
not a matched palette effect. The last analysis compares feedback with choosing the best of $k$
one-shot runs by support-start success (ties going to the first run) at the same token cost, where $k$ is the rounded ratio of
the mean tokens of a W2 session to those of a W1 session. At matched output tokens the ratio is
2.1--3.5, so $k=2$ for Astra and Sol 6.1 and $k=3$ for Opus 5.5 and Terra, with best-of-2 averaged
over the three pairs of runs. Best-of-$k$ then reaches 0.48, 0.35, 0.24 and 0.27, against 0.56,
0.43, 0.36 and 0.33 for W2. At matched total tokens (input and output) the ratio is 3.4--5.1, so
$k$ would exceed the three one-shot runs per family except for Astra. Best-of-3, which uses
59--88\% of W2's total tokens, gives 0.51, 0.37, 0.24 and 0.27.

\begin{table}[h]
\centering\small\setlength{\tabcolsep}{3.8pt}
\caption{Where the FreeArch gain comes from (held-out-level success with feedback, family macro over the
20 private families). A: the fixed actor's graph in the FreeArch language, only weights free. B: any
graph with the fixed actor's operations. Differences are paired over families with bootstrap 95\%
CIs, and pooled rows average the four writers per family first.}
\label{tab:fa_controls}
\begin{tabular}{lrrrrlll}
\toprule
Writer & Fixed & A & B & FreeArch & A$-$fixed & B$-$A & FreeArch$-$B\\
\midrule
GPT-6 Astra & 0.56 & 0.58 & 0.88 & 0.90 & +0.02 [-0.03,+0.08] & +0.30 [+0.17,+0.43] & +0.02 [-0.02,+0.06]\\
GPT-6.1 Sol & 0.43 & 0.44 & 0.78 & 0.84 & +0.00 [-0.04,+0.04] & +0.35 [+0.24,+0.46] & +0.06 [-0.03,+0.15]\\
Claude Opus 5.5 & 0.36 & 0.35 & 0.56 & 0.68 & -0.01 [-0.08,+0.06] & +0.22 [+0.12,+0.32] & +0.12 [+0.05,+0.21]\\
GPT-6.1 Terra & 0.33 & 0.37 & 0.68 & 0.74 & +0.03 [-0.00,+0.07] & +0.31 [+0.22,+0.41] & +0.06 [-0.00,+0.13]\\
\midrule
Pooled (4 writers) & 0.42 & 0.43 & 0.73 & 0.79 & +0.01 [-0.02,+0.04] & +0.29 [+0.21,+0.37] & +0.06 [+0.02,+0.10]\\
\bottomrule
\end{tabular}
\end{table}

\begin{table}[h]
\centering\small
\caption{Weights in the writer's own FreeArch graph (Astra, FreeArch W2 run 0, one graph per private family):
written versus learned from re-initialised weights, with the same learners in the fixed actor for
reference (seed 0). Family macro; diverged runs score 0; the difference is paired over families
with a bootstrap 95\% CI.}
\label{tab:learned_fa}
\begin{tabular}{lrrl}
\toprule
Weights in the writer's FreeArch graph & S-start & Q-lvl & Written $-$ learned [95\% CI]\\
\midrule
Written (Astra, FreeArch W2 run 0) & 1.00 & 0.89 & \\
Learned: BC-support & 0.01 & 0.01 & +0.89 [+0.77,+0.98] (19/20)\\
Learned: distill (no aug.) & 0.01 & 0.01 & +0.89 [+0.77,+0.98] (19/20)\\
\midrule
Fixed actor: BC-support & 0.96 & 0.12 & \\
Fixed actor: distill (no aug.) & 0.90 & 0.09 & \\
\bottomrule
\end{tabular}
\end{table}

\begin{table}[h]
\centering\footnotesize\setlength{\tabcolsep}{4pt}
\caption{Protocol-specified secondary analyses (private tier, main writers, with feedback,
family macro unless stated). Deliv.: fraction of the 60 sessions with an artifact. Q-lvl: primary,
undelivered sessions score 0. Cond.\ deliv.: mean over delivered sessions. Excl.\ undeliv.: family
means over delivered runs only (number of families in parentheses where one drops out). Q-pal:
held-out levels with permuted palettes, level starts only, on the 19 families where the palette
changes the frame, with undelivered sessions scored 0. Last column: runs with Q-lvl $\ge0.5$, with a
Wilson 95\% interval.}
\label{tab:prespec}
\begin{tabular}{llrrrrrl}
\toprule
Arm & Writer & Deliv. & Q-lvl & Cond. deliv. & Excl. undeliv. & Q-pal & runs $\ge0.5$ [Wilson 95\% CI]\\
\midrule
W2 & GPT-6 Astra & 1.00 & 0.56 & 0.56 & 0.56 & 0.54 & 31/60 [0.39, 0.64]\\
 & GPT-6.1 Sol & 1.00 & 0.43 & 0.43 & 0.43 & 0.39 & 25/60 [0.30, 0.54]\\
 & Claude Opus 5.5 & 0.87 & 0.36 & 0.41 & 0.39 (19) & 0.31 & 18/60 [0.20, 0.43]\\
 & GPT-6.1 Terra & 1.00 & 0.33 & 0.33 & 0.33 & 0.30 & 21/60 [0.24, 0.48]\\
\midrule
P2 & GPT-6 Astra & 1.00 & 1.00 & 1.00 & 1.00 & 1.00 & 60/60 [0.94, 1.00]\\
 & GPT-6.1 Sol & 1.00 & 1.00 & 1.00 & 1.00 & 1.00 & 60/60 [0.94, 1.00]\\
 & Claude Opus 5.5 & 1.00 & 0.97 & 0.97 & 0.97 & 0.97 & 59/60 [0.91, 1.00]\\
 & GPT-6.1 Terra & 1.00 & 0.98 & 0.98 & 0.98 & 0.99 & 60/60 [0.94, 1.00]\\
\midrule
FreeArch W2 & GPT-6 Astra & 1.00 & 0.90 & 0.90 & 0.90 & 0.90 & 56/60 [0.84, 0.97]\\
 & GPT-6.1 Sol & 1.00 & 0.84 & 0.84 & 0.84 & 0.83 & 53/60 [0.78, 0.94]\\
 & Claude Opus 5.5 & 0.98 & 0.68 & 0.70 & 0.69 & 0.66 & 44/60 [0.61, 0.83]\\
 & GPT-6.1 Terra & 1.00 & 0.74 & 0.74 & 0.74 & 0.73 & 49/60 [0.70, 0.89]\\
\bottomrule
\end{tabular}
\end{table}

\begin{figure}[!htbp]
\centering
\includegraphics[width=\linewidth]{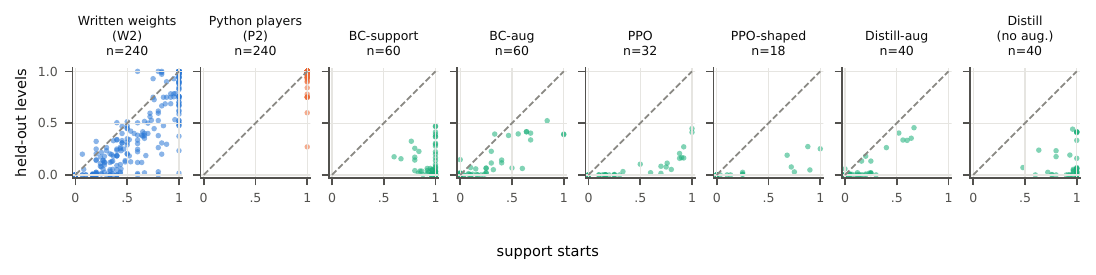}
\caption{Support-start success (x) against held-out-level success (y) per artifact ($n$: one per
writer run, distilled P2 run, or training seed and family, with the 32 sparse-reward runs of
Table~\ref{tab:baselines} for PPO). Unaugmented learners fit the support set and collapse on held-out levels, while augmented learners
fit neither. Written weights lie closer to the diagonal. Python players cluster near $(1,1)$.}
\label{fig:support}
\end{figure}

\begin{table}[h]
\centering\footnotesize
\caption{Q-lvl by difficulty stratum (number of families in parentheses, family macro,
main writers pooled). For W2 and P2, brackets give 95\% bootstrap CIs over the families within
a stratum (none for the one-family stratum). ``Best learned'' takes, per family, the best of
BC-support, BC-aug, dihedral-only BC, both distillation variants and PPO. BC-oracle trains on states
of held-out levels and only shows the actor's capacity.}
\label{tab:strata}
\begin{tabular}{lrrrrr}
\toprule
Policy source & easy (3) & medium (11) & hard (5) & very hard (1) & macro\\
\midrule
Written weights, W2 & 0.89 {\scriptsize[0.79,0.99]} & 0.40 {\scriptsize[0.27,0.55]} & 0.27 {\scriptsize[0.08,0.48]} & 0.00 & 0.42\\
Written weights, W1 & 0.65 & 0.25 & 0.14 & 0.00 & 0.27\\
Python player, P2 & 0.98 {\scriptsize[0.96,1.00]} & 0.99 {\scriptsize[0.97,1.00]} & 0.99 {\scriptsize[0.97,1.00]} & 0.96 & 0.99\\
Python player, P1 & 0.97 & 0.93 & 0.94 & 0.84 & 0.93\\
Reactive Python, P-react & 0.42 & 0.44 & 0.38 & 0.08 & 0.40\\
Best learned baseline (non-privileged) & 0.32 & 0.16 & 0.05 & 0.00 & 0.15\\
BC-oracle (capacity) & 0.96 & 0.89 & 0.89 & 0.83 & 0.90\\
\bottomrule
\end{tabular}
\end{table}

\begin{table}[h]
\centering\footnotesize
\caption{Without the three preliminary-study families (\texttt{ztl}, \texttt{lys}, \texttt{esm}):
held-out-level success with feedback on the other 17 private families (family macro). Upper block:
main writers pooled, by stratum, with the value on all 20 families in parentheses where it
differs. Lower block: per writer, with the paired FreeArch W2$-$W2 difference (bootstrap 95\% CI over
families, with the number of families where FreeArch is better).}
\label{tab:prelim}
\begin{tabular}{lrrrrr}
\toprule
Main writers pooled & easy (2) & medium (9) & hard (5) & very hard (1) & macro\\
\midrule
W2 & 0.89 & 0.31 (0.40) & 0.27 & 0.00 & 0.35 (0.42)\\
FreeArch W2 & 0.92 (0.95) & 0.75 (0.78) & 0.76 & 0.64 & 0.77 (0.79)\\
P2 & 0.98 & 0.99 & 0.99 & 0.96 & 0.98 (0.99)\\
\midrule
Per writer & W2 & FreeArch W2 & P2 & \multicolumn{2}{l}{FreeArch W2 $-$ W2 [95\% CI]}\\
\midrule
GPT-6 Astra & 0.50 & 0.89 & 1.00 & \multicolumn{2}{l}{+0.39 [+0.25,+0.55] (15/17)}\\
GPT-6.1 Sol & 0.35 & 0.82 & 0.99 & \multicolumn{2}{l}{+0.47 [+0.30,+0.63] (15/17)}\\
Claude Opus 5.5 & 0.27 & 0.65 & 0.96 & \multicolumn{2}{l}{+0.37 [+0.25,+0.51] (16/17)}\\
GPT-6.1 Terra & 0.27 & 0.71 & 0.98 & \multicolumn{2}{l}{+0.44 [+0.32,+0.56] (16/17)}\\
\bottomrule
\end{tabular}
\end{table}

\begin{table}[h]
\centering\footnotesize
\caption{Ablations on six stratified families (\texttt{ztl}, \texttt{lys}, \texttt{esm},
\texttt{ruz}, \texttt{crg}, \texttt{mds}), with one run each except for the main W1 and P1, which average
three runs. M+S: the source includes solver code. No axial: the actor without row and column
broadcasts. 8 iter.: 8 instead of 128 tied iterations. Undelivered sessions score 0: Opus 5.5
delivered 5, 5, 3 and 4 of 6 sessions for W1 M+S, no axial, 8 iter.\ and effort med., Sol 6.1 5 of 6
without axial broadcasts, and Astra and Sol 6.1 5 of 6 for P1 M+S.}
\label{tab:ablations}
\begin{tabular}{lrrrrrrr}
\toprule
Writer & W1 & W1, M+S & no axial & 8 iter. & effort med. & P1 & P1, M+S\\
\midrule
GPT-6 Astra & 0.43 & 0.53 & 0.50 & 0.16 & 0.39 & 0.94 & 0.73\\
GPT-6.1 Sol & 0.39 & 0.41 & 0.33 & 0.28 & 0.29 & 0.96 & 0.73\\
Claude Opus 5.5 & 0.15 & 0.21 & 0.13 & 0.23 & 0.18 & 0.89 & 0.84\\
\bottomrule
\end{tabular}
\end{table}

\begin{table}[h]
\centering\footnotesize
\caption{Closed-loop controls for written actors (W2, family macro over the private tier).
Zero frame: the observation is replaced by a constant frame. First frame: frozen at the
episode's first frame. Perturbed, no RESET: success with RESET disabled on the perturbed starts of
held-out and support levels (Q-lvl and Q-off, pooled). CF
sensitivity: the fraction of counterfactual frames (a frame from another episode, inserted at a
random step) that change the chosen action. The last row gives the floors.}
\label{tab:controls}
\begin{tabular}{lrrrrr}
\toprule
Writer (W2) & Q-lvl & zero frame & first frame & perturbed, no RESET & CF sensitivity\\
\midrule
GPT-6 Astra & 0.56 & 0.00 & 0.07 & 0.65 & 0.75\\
GPT-6.1 Sol & 0.43 & 0.00 & 0.05 & 0.48 & 0.76\\
Claude Opus 5.5 & 0.36 & 0.02 & 0.05 & 0.45 & 0.75\\
GPT-6.1 Terra & 0.33 & 0.00 & 0.06 & 0.41 & 0.71\\
\midrule
Replay floor / best open loop & 0.02 / 0.04 & & & & \\
\bottomrule
\end{tabular}
\end{table}

\begin{table}[h]
\centering\footnotesize
\caption{Public tier (25 official ARC-AGI-3 games, possibly seen, held-out levels with one level start and up to 20
perturbed starts each). Family macro with bootstrap 95\% CI.}
\label{tab:public}
\begin{tabular}{lllll}
\toprule
Writer & W1 & W2 & P1 & P2\\
\midrule
GPT-6 Astra & 0.06 {\scriptsize[0.01,0.14]} & 0.08 {\scriptsize[0.01,0.18]} & 0.59 {\scriptsize[0.47,0.72]} & 0.75 {\scriptsize[0.64,0.85]}\\
GPT-6.1 Sol & 0.04 {\scriptsize[0.00,0.10]} & 0.05 {\scriptsize[0.01,0.12]} & 0.58 {\scriptsize[0.46,0.70]} & 0.68 {\scriptsize[0.55,0.79]}\\
Claude Opus 5.5 & 0.03 {\scriptsize[0.00,0.10]} & 0.06 {\scriptsize[0.00,0.15]} & 0.31 {\scriptsize[0.17,0.45]} & 0.43 {\scriptsize[0.29,0.56]}\\
GPT-6.1 Terra & 0.01 {\scriptsize[0.00,0.03]} & 0.02 {\scriptsize[0.00,0.05]} & 0.51 {\scriptsize[0.39,0.64]} & 0.57 {\scriptsize[0.44,0.69]}\\
GPT-6 Luna & 0.00 {\scriptsize[0.00,0.02]} & -- & 0.06 {\scriptsize[0.01,0.12]} & --\\
Claude Sonnet 5.5 & 0.00 {\scriptsize[0.00,0.01]} & -- & 0.18 {\scriptsize[0.08,0.30]} & --\\
Claude Haiku 5.5 & 0.00 {\scriptsize[0.00,0.00]} & -- & 0.05 {\scriptsize[0.00,0.12]} & --\\
\bottomrule
\end{tabular}
\end{table}

\clearpage
\makeatletter
\setlength{\@fptop}{0pt}
\setlength{\@fpsep}{18pt}
\setlength{\@fpbot}{0pt plus 1fil}
\makeatother

\begin{table}[h]
\centering\footnotesize
\caption{Public tier with feedback: fixed actor, free architecture and Python players
(held-out-level success, mean over 25 games, with the paired FreeArch$-$fixed difference with bootstrap 95\% CI
over games).}
\label{tab:fa_public}
\begin{tabular}{lrrrl}
\toprule
Writer & W2 fixed & W2 FreeArch & P2 & $\Delta$ FreeArch$-$fixed [95\% CI]\\
\midrule
GPT-6 Astra & 0.08 & 0.42 & 0.75 & +0.34 [+0.24,+0.44] (25 games)\\
GPT-6.1 Sol & 0.05 & 0.36 & 0.68 & +0.31 [+0.20,+0.42] (25 games)\\
Claude Opus 5.5 & 0.06 & 0.13 & 0.43 & +0.07 [+0.01,+0.16] (25 games)\\
GPT-6.1 Terra & 0.02 & 0.26 & 0.57 & +0.24 [+0.14,+0.36] (25 games)\\
\bottomrule
\end{tabular}
\end{table}

\begin{table}[h]
\centering\footnotesize
\caption{Size and warmed-up latency per step (median [90th percentile], exact FP32, one idle H200
GPU and one CPU thread). MACs and node evaluations follow the FreeArch counting rule
(Section~\ref{sec:fa-setup}). $^\dagger$ CUDA-graph capture failed for the fixed actor, so its eager time is shown
(few large kernels, so little overhead to remove). $^\ddagger$ sandboxed \texttt{act()} wall time,
as evaluated.}
\label{tab:latency}
\begin{tabular}{lrrr}
\toprule
Measure (median [p90]) & Free architecture & Fixed actor & Python player\\
\midrule
Parameters & 30,157 [99,383] & 57,289 & --\\
GMAC per step & 0.32 [1.23] & 1.33 & --\\
Node evaluations per step & 2,362 [9,671] & 412 & --\\
GPU, batch 1, eager (ms) & 23.8 [115.3] & 4.8 & --\\
GPU, batch 1, CUDA graph (ms) & 4.6 [22.0] & 4.8$^\dagger$ & --\\
GPU, batch 64, per action ($\mu$s) & 525 [2065] & 273 & --\\
CPU, 1 thread, batch 1 (ms) & 35.8 [312] & 36.3 & 0.82 [2.78]$^\ddagger$\\
\bottomrule
\end{tabular}
\end{table}

\end{document}